\documentclass[a4paper,fleqn]{cas-dc}

\usepackage[numbers]{natbib}
\usepackage{tikz}
\usepackage{hyperref}       
\usepackage{soul}
\usepackage{booktabs}
\usepackage{algorithm}
\usepackage{algorithmic}

\def\tsc#1{\csdef{#1}{\textsc{\lowercase{#1}}\xspace}}
\tsc{WGM}
\tsc{QE}
\tsc{EP}
\tsc{PMS}
\tsc{BEC}
\tsc{DE}

\newcommand*\diff{\mathop{}\!\mathrm{d}}

\begin{document}
\let\WriteBookmarks\relax
\def\floatpagepagefraction{1}
\def\textpagefraction{.001}
\shorttitle{Physics-informed neural particle flow}
\shortauthors{D. Csuzdi et~al.}

\title [mode = title]{Physics-informed neural particle flow for the Bayesian update step}                      
\tnotemark[1]

\tnotetext[1]{The project supported by the Doctoral Excellence Fellowship Programme (DCEP) is funded by the National
Research Development and Innovation Fund of the Ministry of Culture and Innovation and the Budapest
University of Technology and Economics.\\This work was funded by the National Research, Development and Innovation Office (NKFIH) under NKKP Grant Agreement No. ADVANCED 152880.}

\author[1]{Domonkos Csuzdi}[type=editor,
                        auid=000,bioid=1,
                        orcid=0000-0003-4774-3330]

\ead{domonkos.csuzdi@edu.bme.hu}

\credit{Conceptualization, Methodology, Software, Visualization, Writing - original draft, Writing - review \& editing}

\affiliation[1]{organization={Department of Control for Transportation and Vehicle Systems, Faculty of Transportation Engineering and Vehicle Engineering, Budapest University of Technology and Economics},
                addressline={Műegyetem rkp.~3}, 
                city={Budapest},
                postcode={H-1111}, 
                country={Hungary}}

\author[1]{Tamás Bécsi}[ orcid=0000-0002-1487-9672]
\ead{becsi.tamas@kjk.bme.hu}
\credit{Funding acquisition, Project administration, Supervision, Writing - original draft, Writing - review \& editing}

\author[1]{Olivér Törő}[orcid=0000-0002-7288-5229]

\cormark[1]
\ead{toro.oliver@kjk.bme.hu}

\credit{Conceptualization, Formal analysis, Methodology, Validation, Writing - original draft, Writing - review \& editing}

\cortext[cor1]{Corresponding author}

\begin{abstract}
The Bayesian update step poses significant computational challenges in high-dimensional nonlinear estimation. While log-homotopy particle flow filters offer an alternative to stochastic sampling, existing formulations usually yield stiff differential equations. Conversely, existing deep learning approximations typically treat the update as a black-box task or rely on asymptotic relaxation, neglecting the exact geometric structure of the finite-horizon probability transport. In this work, we propose a physics-informed neural particle flow, which is an amortized inference framework.
To construct the flow, we couple the log-homotopy trajectory of the prior to posterior density function with the continuity equation describing the density evolution. This derivation yields a governing partial differential equation (PDE), referred to as the master PDE. By embedding this PDE as a physical constraint into the loss function, we train a neural network to approximate the transport velocity field. This approach enables purely unsupervised training, eliminating the need for ground-truth posterior samples. We demonstrate that the neural parameterization acts as an implicit regularizer, mitigating the numerical stiffness inherent to analytic flows and reducing online computational complexity. Experimental validation on multimodal benchmarks and a challenging nonlinear scenario confirms better mode coverage and robustness compared to state-of-the-art baselines.
\end{abstract}

\begin{graphicalabstract}
\includegraphics[width=\textwidth]{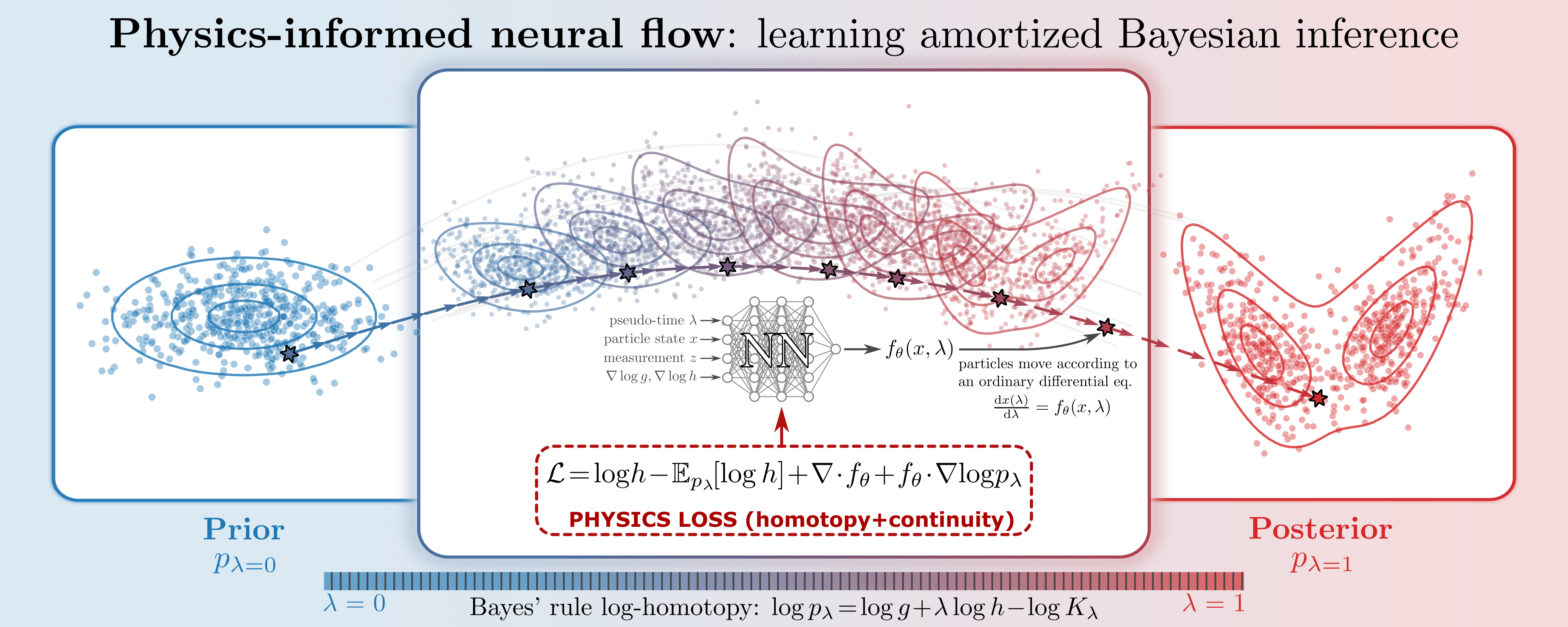}
\end{graphicalabstract}

\begin{highlights}
\item Bayesian update via physics-informed deterministic particle flow
\item Coupling the log-homotopy with the continuity equation yields a physical constraint
\item Neural parameterization reduces the numerical stiffness of exact analytical flows
\item The learned amortized operator generalizes to out-of-distribution priors
\end{highlights}


\maketitle

\section{Introduction}

At its core, Bayesian inference is a problem of integration. Computing the posterior expectation of the state requires normalizing the product of a prior and a likelihood. This task involves evaluating high-dimensional integrals that are typically not available in closed form. To overcome this intractability, computational statistics use Markov chain Monte Carlo (MCMC) methods \cite{chen2012monte,jones2022markov}, such as the Metropolis--Hastings algorithm. These methods construct a Markov chain that has the target density as its stationary distribution. This way, complex integrals can be evaluated as ergodic averages \cite{asmussen2011new}.

In the context of dynamic state estimation, where the probability density function (pdf) must be updated recursively as new data arrive, sequential Monte Carlo (SMC) estimators \cite{barbu_monte_2020,robert2004monte}, also known as particle filters (PFs), have become a standard approximation technique \cite{doucet_sequential_2000,doucet_sequential_2001,arulampalm2002PF}. SMC extends the Monte Carlo approach to dynamic systems by representing the evolving pdf with a set of weighted samples. However, both MCMC and SMC face severe limitations in high-dimensional state spaces. Standard sampling methods suffer from the curse of dimensionality, leading to slow mixing in MCMC chains and weight degeneracy in particle filters.

The theoretical connection between statistical mechanics and multivariate optimization, established by Kirkpatrick et al. \cite{kirkpatrick1983optimization}, provided the basis for simulated annealing and motivated the application of stochastic differential equations, such as Langevin dynamics, to probabilistic inference. By treating the probability density as a physical system relaxing toward thermal equilibrium, these methods apply stochastic forces to avoid getting stuck in local optima and explore the state space more effectively than simple random walks.

Although stochastic approaches guarantee asymptotic convergence, they are limited by the diffusive nature of the exploration process. Random walks can be computationally inefficient, particularly in high-dimensional spaces. To transition from the prior to the posterior, particles must traverse regions of low probability. To address this issue, particle flow filters have emerged as an alternative. In contrast to SMC methods, these algorithms transport samples along a continuous trajectory.

An important realization of this approach is the log-homotopy particle flow filter, introduced by Daum and Huang~\cite{daum2007nonlinear, daum2010exact}. The core principle is to define a continuous deformation of the probability density function, parameterized by a pseudo-time variable $\lambda \in [0, 1]$, which smoothly bridges the prior distribution and the posterior. The particle trajectories are then either governed by a deterministic ordinary differential equation (ODE) or a stochastic differential equation (SDE) derived to match the evolution of this density. However, deriving a valid transport map presents a significant challenge \cite{daum2010exact17,dai2021new}. Analytical solutions derived from the log-homotopy often yield numerically stiff differential equations that pose stability problems for standard integrators \cite{daum2014seven,dai2022role,dai2022stiffness,crouse2021particle}.

Conversely, deep learning-based approximations offer flexibility but frequently treat the update as a black box. Notable examples include normalizing flows \cite{rezende2015variational, papamakarios2021normalizing} and conditional neural processes \cite{garnelo2018conditional}, which parameterize the transport map using invertible neural networks. These methods typically rely on optimizing a statistical divergence such as the Kullback--Leibler (KL) divergence between the transported prior and the target posterior. While effective for general density estimation, this formulation ignores the specific geometric structure of the Bayesian update. Recent approaches that explicitly model particle dynamics, such as the particle flow version of Bayes' rule \cite{chen2019particle}, rely on meta-learning strategies applied to synthetic tasks and assume that convergence to the posterior occurs asymptotically. However, these methods do not enforce the finite-time continuity equation that is derived from the log-homotopy.

This paper focuses on the discrete-time Bayesian update step, isolating it from the prediction dynamics. This separation allows us to examine the fusion of the prior and the likelihood, independent of any particular temporal dynamics or transition models. Under the Markov assumption, the prior distribution contains all historical information, allowing the multiplication and normalization operations to be addressed in a static framework without loss of generality.

In this work, we propose a novel framework that addresses the limitations of both traditional sampling and modern flow-based methods. We introduce a physics-informed neural particle flow (PINPF), which implements the Bayesian update via deterministic particle transport driven by a neural network. 
Coupling the continuity equation with the log-homotopy that interpolates between the prior and the posterior, results in an implicit partial differential equation (PDE) for the velocity field.
Since this equation is underdetermined and admits a broad family of valid transport solutions, it serves as a general governing law, which we call therefore the master PDE. Unlike black-box neural operators, our method explicitly uses this PDE to constrain the learning process. By embedding this physical law into training, we obtain an amortized inference operator trained in a fully unsupervised manner. This approach combines the theoretical rigor of analytical particle flows with the flexibility of deep learning. Neural parameterization reduces numerical stiffness associated with exact flow filters while maintaining a computational complexity of $O(N)$ per update. The implementation of our algorithm can be accessed at \url{https://github.com/DomonkosCs/PINPF}.

\subsection{Notations}
The real-valued $n$-dimensional Euclidean space is denoted by $\mathbb{R}^n$.
For a scalar function $f(x) : \mathbb{R}^n \rightarrow \mathbb{R}$, its gradient is
$\nabla f(x) = [\partial f / \partial x_1, \partial f / \partial x_2, \dots, \partial f / \partial x_n]^\top \in \mathbb{R}^n$.
For a vector field $v(x) : \mathbb{R}^n \rightarrow \mathbb{R}^n$, its divergence is
$\nabla \cdot v(x) = \sum_{i=1}^n \partial v_i / \partial x_i \in \mathbb{R}$.
Unless explicitly stated otherwise, the operator $\nabla$ refers to differentiation with respect to $x$.
The Laplace operator $\Delta$ is defined by $\Delta f(x) = \sum_{i=1}^n \partial^2 f / \partial x_i^2 \in \mathbb{R}$.

We adopt the following simplified notation for the one-step Bayesian inference throughout this paper.
Let $x \in \mathbb{R}^{n_x}$ denote the random state vector and $z \in \mathbb{R}^{n_z}$ the random measurement vector in the $n_x$- and $n_z$-dimensional spaces, respectively.
Let $p_x(x)$ denote the prior probability density function of $x$ and $p_z(z\mid x)$ the likelihood of a measurement $z$.
According to Bayes' theorem, the posterior conditional density function of $x$ given a measurement $z$ is
\begin{equation}
    p_x(x|z) = \frac{p_x(x)p_z(z|x)}{p_z(z)} \, ,
\end{equation}
where the normalizing constant is $p_z(z) = \int p_x(x) p_z(z|x) \diff x$.
We assume that all probability density functions exist, are sufficiently differentiable, and are nonvanishing everywhere.
For simplicity, we use the abbreviations $p(x) = p_x(x|z)$, $g(x) = p_x(x)$, and $h(x) = p_z(z|x)$.

Discrete time steps are indexed by $k$, while $t\in \mathbb{R}$ denotes a continuous flow parameter and $\lambda \in [0,1]$ is the continuous homotopy parameter.
Probability density functions parameterized by $t$ or $\lambda$ are denoted by $p_t$ or $p_\lambda$, respectively.

The expectation w.r.t. the density $p$ is denoted by $\mathbb{E}_p$.

\subsection{Plan of the paper}

The remainder of the paper is organized as follows. Section \ref{sec:sota} reviews the relevant algorithms for implementing the Bayesian update step. Techniques are categorized into sampling-based, transport-based, and variational approaches. Section \ref{sec:lhpf} establishes the mathematical foundation of the proposed method and introduces the master PDE for the general deterministic log-homotopy particle flow.

In Section \ref{sec:theo}, we perform a theoretical analysis of the master PDE. We verify its consistency with the global conservation of probability mass and derive its explicit solution in one dimension. To give a broader perspective, we examine the theoretical links connecting stochastic relaxation and deterministic transport methods and provide a comparative analysis of the discussed algorithms.

Section \ref{sec:pinpf} details the proposed PINPF framework. We formulate the physics-informed loss function based on the residual of the master PDE, describe the feature construction strategy, and present the sequential local learning algorithm. Section \ref{sec:exp} evaluates the method on a four-dimensional multimodal Gaussian mixture problem and a challenging nonlinear time-difference-of-arrival tracking scenario, demonstrating its ability to mitigate numerical stiffness and generalize to out-of-distribution priors. Section \ref{sec:conc} concludes the paper, and \ref{sec:app} presents additional qualitative results and visualizations of the training data.

\section{Implementing the Bayesian update} \label{sec:sota}

In the Bayesian framework, Bayes’ theorem serves as a fundamental tool for probabilistic inference~\cite{jaynes2003probability,sarkka_bayesian_2023,taghvaei2024implement}. In practice, the posterior distribution is often estimated by pointwise multiplication of the prior pdf and the likelihood function. For specific distributions, such as those from the exponential family, conjugate priors can be used, yielding a closed-form solution in which the posterior remains in the same family~\cite[Chapter~3]{raiffa2000applied}. When such analytical convenience is unavailable, numerical grid-based methods such as the histogram filter can perform the multiplication explicitly on a discretized grid~\cite[Chapter~4]{thrun2005probabilistic}. However, the resulting function does not automatically integrate to unity. To ensure that the posterior describes a valid probability distribution, normalization is required, which requires computing an integral over the entire space. This problem is generally intractable, especially in high dimensions. 

Efficient algorithms aim to avoid this computationally intensive integration. The most well-known specialized estimator is the Kalman filter, which operates in the linear-Gaussian regime and provides an exact analytical solution where normalization is implicit~\cite{kalman1960,chen2003bayesian}. Regarding numerical approaches, the particle filter approximates the posterior with a finite set of weighted samples, effectively reducing the normalization problem to the normalization of scalar weights~\cite{gordon1993novel}.

It should be noted that this distinct update step is an artifact of discrete-time formulations. In continuous-time systems, the sequential distinction between prediction and update disappears. Instead, the posterior distribution evolves simultaneously under the influence of system dynamics and incoming observations. Mathematically, the evolution of the probability density function is governed by the Kushner--Stratonovich equation~\cite{stratonovich1960conditional,kushner1964conditional}. In this stochastic differential equation, the Bayesian update appears not as a pointwise multiplication but as an additive innovation term driven by the discrepancy between observed and expected measurements. In the specific case of linear dynamics and Gaussian noise, this infinite-dimensional problem simplifies to a finite set of ordinary differential equations known as the Kalman--Bucy filter~\cite{kalman1961new}. The connection between discrete- and continuous-time filtering is discussed in detail in \cite{jazwinski2007stochastic} and \cite{sarkka2019applied}, covering both discrete-to-continuous and continuous-to-discrete approaches.

\subsection{Bayesian update in sequential estimation}

In recursive estimation tasks arising in state-space systems, the general intractability of the update step has driven a significant body of research toward SMC methods~\cite{barbu_monte_2020,robert2004monte}. Particle filters approximate the posterior pdf $p_k(x)$ at time step $k$ using a finite set of weighted random samples (particles) $x_k^{i}$ with associated importance weights $w_k^{i} \ge 0$ $(i=1,\dots,N)$. The empirical pdf is given by the Dirac mixture
\begin{equation} \label{eq:dirac_approx}
    p_k(x) \approx \sum_{i=1}^N w_k^{i} \delta(x - x_k^{i}) \, ,
\end{equation}
where $\delta(\cdot)$ denotes the Dirac delta. For \eqref{eq:dirac_approx} to represent a valid pdf, the weights must sum to unity.

In the particle filter framework~\cite{gordon1993novel}, particles are propagated through the system dynamics, and their weights are updated according to the likelihood of the observed data. However, standard SMC algorithms suffer from inherent degeneracies~\cite{arulampalm2002PF,doucet_sequential_2000,doucet_sequential_2001,bishop2006pattern}. A well-known phenomenon is weight degeneracy: after a few iterations, the variance of the weights can increase uncontrollably, leading to a situation in which all but one particle carry negligible weight. To counteract this, a resampling step is typically introduced to discard particles with low weights and duplicate those with high weights. While resampling preserves the statistical validity of the filter, it introduces a new issue known as particle depletion or sample impoverishment. This results in a loss of diversity, with the particle cloud collapsing to a single point in state space and failing to represent posterior uncertainty adequately. Furthermore, these methods are susceptible to the curse of dimensionality~\cite{daum2003curse}. In high-dimensional state spaces, the volume of the region where the likelihood is significant becomes exponentially small relative to the prior support. It has been shown that the number of particles required to avoid weight collapse scales exponentially with the state dimension~\cite{bengtsson2008curse,snyder2008obstacles,rebeschini2015can}. Interestingly, a large number of independent observations can also lead to filter collapse~\cite{van_leeuwen_nonlinear_2015}.

To overcome particle depletion and weight collapse caused by assimilating highly informative observations, various progressive Bayesian update methods have been proposed. The underlying idea is to avoid pointwise multiplication of the prior by the full likelihood function in a single step. Instead, measurement assimilation is decomposed into a sequence of partial updates that are numerically stable.

An early approach of this kind was introduced by Oudjane and Musso for regularized particle filters~\cite{oudjane2000progressive}. Their method factorizes the likelihood function in a principled way into smaller sub-likelihoods and introduces a decreasing sequence of covariance matrices for the observation noise.  A similar method for Gaussian problems is presented in \cite{hanebeck2013pgf}.

Another progressive update method for Bayesian state estimation was proposed by Hanebeck~\cite{hanebeck2003progressive}. In this approach, the probability density evolves according to a system of linear first-order ODEs defined over an artificial progression parameter ranging from zero to one. The ODEs are derived by minimizing the squared integral deviation between the true and the approximated density, which governs the evolution of the density parameters during the update.

\subsection{Sampling methods}

In the Bayesian framework, a significant computational challenge in high-dimensional inference is sampling from the posterior and evaluating expectations. According to Bayes' theorem, the posterior density $p(x)$ is given by
\begin{equation}
p(x) = \frac{h(x) g(x)}{\int h(x) g(x) \diff x} \, ,
\end{equation}
where $h$ is the likelihood and $g$ is the prior pdf. One can express the posterior in Boltzmann--Gibbs form as
\begin{equation}
    p(x) = \frac{1}{K} e^{-V(x)} \, ,
\end{equation}
where $V(x) = -\log h(x) - \log g(x)$ is the potential energy function, and $K = \int e^{-V(x)} \diff x$ is the intractable normalizing constant. Because $K$ is generally unavailable in closed form, direct sampling is impossible for most nontrivial models.

To overcome this, MCMC methods construct an ergodic Markov chain $x^{(j)}$, $(j = 1 , \dots , J)$ that admits $p$ as its unique stationary distribution. Under weak regularity conditions, the ergodic theorem guarantees that sample averages converge to the desired posterior expectations~\cite{tierney1994markov}:
\begin{equation}
    \lim_{J \to \infty} \frac{1}{J} \sum_{j=1}^J \varphi(x^{(j)}) = \mathbb{E}_{p}[\varphi(x)] \, ,
\end{equation}
where $\varphi$ is an integrable function of the state.

The most general algorithm in this class is the Metropolis--Hastings algorithm~\cite{metropolis1953equation,hastings1970monte}. At each step $j$, a candidate state $x'$ is sampled from a user-defined proposal distribution $q(x'\mid x^{(j)})$, such as a Gaussian random walk centered at the current state $x^{(j)}$. The candidate is then accepted with probability
\begin{equation}
    \min\left(1, \frac{p(x')q(x^{(j)}\mid x')}{p(x^{(j)})q(x'\mid x^{(j)})}\right)\, .
\end{equation}
Since the target density $p(x) = e^{-V(x)}/K$ appears in both the numerator and the denominator, the intractable constant $K$ cancels out, allowing the transition to be computed using the unnormalized pdf.

However, standard random-walk MCMC methods suffer from severe inefficiencies in high dimensions due to the curse of dimensionality \cite{bellman1961adaptive}. Isotropic exploration of the state space results in diffusive behavior, leading to long mixing times and highly correlated samples.

Despite these advances, gradient-based MCMC methods often fail in the presence of multimodal distributions because the Markov chains tend to become trapped in local minima of the potential energy landscape. To overcome this issue, annealed MCMC methods construct a sequence of intermediate distributions that bridge a tractable prior and a complex posterior \cite{neal2001annealed}. A geometric path is typically defined via an inverse-temperature parameter $\beta \in [0,1]$, yielding a family of densities $p_\beta(x) \propto h(x)^\beta g(x)$. This strategy of progressively transforming a simple probability measure into the target distribution has a close theoretical connection to the homotopy-based particle flows discussed in this work, although it is implemented in this context via discrete stochastic transitions rather than continuous deterministic transport.

\subsection{Transport methods and normalizing flows}

While annealed MCMC relies on stochastic transitions to bridge the initial and target distributions, an alternative paradigm is deterministic measure transport. The objective is to construct a diffeomorphism $M: \mathbb{R}^d \to \mathbb{R}^d$ that pushes a tractable reference density $\rho_0$ forward to the generally intractable target posterior $p$.

Formally, one seeks a map $M$ such that the pushforward of the reference density matches the target, denoted by $M_\# \rho_0 = p$. This condition implies that if $u \sim \rho_0$, then $x = M(u)$ is distributed according to $p$. The densities are related by the change-of-variables formula:
\begin{equation}
    p(M(u)) = \rho_0(u) \left| \det \nabla M(u) \right|^{-1}.
\end{equation}
Once $M$ is approximated, independent posterior samples can be generated at negligible cost by pushing reference samples through the map. This effectively converts the integration problem of Bayesian inference into an optimization problem over function spaces.

The existence of such maps is guaranteed by optimal transport theory. In particular, Brenier's theorem~\cite{brenier1991polar,bonnotte2013knothe} states that there exists a unique map that minimizes the quadratic Wasserstein cost and is given by the gradient of a convex scalar potential. Alternatively, the Knothe--Rosenblatt rearrangement~\cite{rosenblatt1952remarks,knothe1957contributions} guarantees the existence of a triangular transport map, which is computationally favorable due to its efficiently computable Jacobian determinant.

In the context of machine learning, these maps are realized as normalizing flows (NF)~\cite{rezende2015variational,papamakarios2021normalizing}. Early approaches parameterized the transport map $M$ as a composition of simple, invertible algebraic layers (e.g., planar or radial flows)~\cite{kobyzev2020normalizing}. An improvement over these elementary transformations is the neural spline flow (NSF) proposed by Durkan et al.~\cite{durkan2019neural}. This architecture replaces simple affine coupling layers with monotonic rational-quadratic splines defined on a discretized grid. By enforcing monotonicity, the transformation remains strictly invertible, while the piecewise definition provides sufficient flexibility to model complex, multimodal densities. The Jacobian of these transformations is diagonal or triangular ensuring that determinant computations remain tractable with linear complexity.

Despite the success of discrete flow architectures, recent developments have shifted toward continuous-time formulations, referred to as continuous normalizing flows (CNF)~\cite{chen2018neural}. In this framework, the discrete sequence of layered transformations is replaced by a continuous integration process. The flow of samples is modeled by a first-order ODE, where the time-dependent velocity field $f_\theta$ is parameterized by a neural network with weights $\theta$:
\begin{equation}
    \frac{\diff x(t)}{\diff t} = f_\theta(x(t), t) \, .
\end{equation}
The transformation $M$ is obtained by numerically integrating this equation from an initial time $t_0$ to a terminal time $t_1$. A distinct advantage of this formulation is efficient computation of the change in probability density. Unlike general discrete flows, which may require computing the determinant of the Jacobian at every step, CNFs rely on the instantaneous change-of-variables formula~\cite{kothe2023review}. This formula states that the rate of change of the log-density is determined by the negative divergence of the vector field $f_\theta$. Consequently, the total change in log-density is given by integrating the divergence of the flow along the trajectory:
\begin{equation}
    \log p(x(t_1)) = \log p(x(t_0)) - \int_{t_0}^{t_1} \nabla \cdot f_\theta \, \diff t \, .
\end{equation}
This approach allows the log-density to be computed using standard ODE solvers by treating it as an auxiliary state variable.

The parameters of the flow are typically learned via variational inference (VI) by minimizing the KL divergence $D_{\text{KL}}(M_\# \rho_0 \,\|\, p)$ between the approximated density and the target posterior $p$. It is important to note that the standard training paradigm for NFs is maximum likelihood estimation (MLE), which minimizes the forward KL divergence $D_{\text{KL}}(p \,\|\, M_\# \rho_0)$. However, MLE is a supervised approach requiring independent samples from the target distribution, which are unavailable in the inference setting. Consequently, one is restricted to minimizing the reverse KL divergence \cite{rezende2015variational}. This objective is known to induce mode-seeking behavior, often causing the flow to underestimate posterior variance or drop modes entirely \cite{blei2017variational,papamakarios2021normalizing}. Furthermore, standard NFs are indifferent to the geometry of the transition: they yield a mapping that minimizes a statistical divergence but lacks physical interpretability.

\subsection{Stein variational gradient descent}

Stein variational gradient descent (SVGD), introduced by Liu and Wang~\cite{liu2016stein}, is a deterministic alternative to classical MCMC sampling. While MCMC approaches rely on stochastic ergodic chains to explore the target posterior $p(x)$, SVGD frames sampling as a variational inference task. The objective is to transport a set of particles $\{x_i\}_{i=1}^N$ to approximate $p(x)$ by minimizing the KL divergence via a gradient flow on a reproducing kernel Hilbert space (RKHS).

The transport map is defined as $M(x) = x + \epsilon \phi(x)$, where $\phi$ is a perturbation direction in the RKHS. By maximizing the decay rate of the KL divergence, a closed-form expression for the optimal velocity field $\phi^*$ is obtained as
\begin{equation}
    \label{eq:stein_gradient}
    \phi^*(x) = \mathbb{E}_{q(y)} \big{[} \underbrace{k(x,y) \nabla_y \log p(y)}_{\text{Driving force}} + \underbrace{\nabla_y k(x,y)}_{\text{Repulsive force}} \big{]} \, ,
\end{equation}
where $q$ is the density of the evolving particle set.
The update rule $x_i \leftarrow x_i + \epsilon \phi^*(x_i)$ can be interpreted as the combination of two competing forces:

\begin{enumerate}
    \item {Driving force:} The first term, $k(x,y)\nabla_y \log p(y)$, is a weighted aggregation of the score functions of the particle ensemble. The kernel $k(x,y)$ acts as a spatial smoothing operator, causing the driving force to be computed as a weighted average of the gradients of the surrounding particles rather than relying solely on the local score function. This mechanism allows particles located in regions with negligible gradients to move toward high-probability regions indicated by neighboring particles.
    
    \item {Repulsive force:} The second term, $\nabla_y k(x,y)$, induces a repulsive force that drives particle $x$ away from $y$ when they are in close proximity. This mechanism allows SVGD to avoid mode collapse, which is typical in standard optimization-based sampling.
\end{enumerate}

A standard choice for the kernel is a radial basis function $k(x,y) = \exp(-\|x-y\|^2 / h)$. The bandwidth parameter $h$ 
is usually given by a dynamic strategy using the median heuristic \cite{garreau2017large} as $h = m^2/ \log N$, where
    \begin{equation}
        m = \operatorname{median}\left( \{ \| x_i - x_j \| : 1 \le i < j \le N \} \right) \, .
    \end{equation}

Theoretically, this evolution constitutes the gradient flow of the KL divergence with respect to the Stein geometry~\cite{liu2017stein}. A key distinction from Langevin dynamics, which corresponds to gradient flow in the Wasserstein metric, is the absence of stochastic noise. Instead of relying on Brownian diffusion to explore the state space, SVGD uses deterministic, nonlocal interactions induced by the kernel. This repulsive interaction typically requires fewer particles to cover the support of the target distribution than stochastic methods.

\subsection{Particle flow Bayes' rule}

A recent advance bridging the gap between deterministic transport maps and sequential Bayesian inference is particle flow Bayes' rule (PFBR), proposed by Chen et al.~\cite{chen2019particle}. While classical transport methods often require solving a computationally intensive optimization problem for each new observation, PFBR aims to learn a flexible, amortized operator that generalizes across priors and observations.

The core idea is to represent the Bayesian update as a continuous-time dynamical system. The transport velocity field $f_\theta$ is parameterized by a deep neural network, evolving particles according to the ODE
\begin{equation}
    \frac{\diff x(t)}{\diff t} = f_\theta(x(t), t, \mathcal{S}), \quad t \in [0, T] \, .
\end{equation}
Here, $t$ is a continuous pseudo-time variable, and $T$ is a terminal time chosen to be sufficiently large to reach equilibrium. The velocity depends on the particle state $x$, time $t$, and the current empirical distribution of the particle set $\mathcal{S} = \{x_{i}\}_{i=1}^N$.

The theoretical foundation of PFBR rests on the correspondence between stochastic diffusion processes and deterministic flows. Consider the overdamped Langevin SDE, which targets the posterior distribution $p(x)$:
\begin{equation}
    \diff x(t) = \nabla \log p(x) \diff t + \sqrt{2} \diff w_t \, ,
\end{equation}
where $w_t$ denotes the Wiener process. The evolution of the intermediate probability density $p_t(x)$ of this process is governed by the Fokker-Planck equation \cite{risken1996fokker}:
\begin{equation}
    \frac{\partial p_t(x)}{\partial t} = -\nabla \cdot (p_t(x) \nabla \log p(x)) + \Delta p_t(x) \, .
\end{equation}
By using the identity $\Delta p_t = \nabla \cdot (p_t \nabla \log p_t)$, the diffusion term can be rearranged to express the evolution as a continuity equation:
\begin{equation}
    \frac{\partial p_t(x)}{\partial t} = -\nabla \cdot \left( p_t(x) \left[ \nabla \log p(x) - \nabla \log p_t(x) \right] \right) \, .
\end{equation}
The equivalent deterministic velocity field is identified as:
\begin{equation}
    \frac{\diff x(t)}{\diff t} = \nabla \log p(x) - \nabla \log p_t(x) \, .
\end{equation}
In the PFBR framework, the network $f_\theta$ is trained via meta-learning to approximate this velocity field. By minimizing the KL divergence over a distribution of synthetic inference tasks, the resulting model acts as a general-purpose Bayesian update operator that transports particles from the prior to the posterior.

\subsubsection{Finite-time vs. asymptotic flow formulations} \label{sec:finiteVSasymp}

Chen et al.~\cite{chen2019particle} established the existence of the transport flow by exploiting the correspondence between the Fokker--Planck equation of a stochastic Langevin process and the continuity equation of a deterministic system. They showed that the marginal density evolution of the stochastic process can be exactly replicated by an ODE. To resolve the dependence of this flow on the intractable instantaneous density, the problem is framed as a deterministic optimal control task. By exploiting the equivalence between closed-loop and open-loop controls, they proved the existence of a fixed velocity field that depends only on the initial distribution.

The analytical structure of velocity fields moving particles along Bayesian log-homotopy trajectories was introduced by Daum and Huang \cite{daum2007nonlinear,daum2010exact}. A key distinctin between log-homotopy flows and the PFBR is the temporal formulation. PFBR relies on the asymptotic convergence $t \to \infty$ of the probability-flow ODE toward equilibrium. 
In contrast, a Daum--Huang type filter uses an ODE or SDE to transport particles over a compact pseudo-time interval $\lambda \in [0,1]$.
PFBR does not explicitly enforce the finite-horizon log-homotopy constraint, which leaves the analytical structure of the Bayesian update implicit. Consequently, the neural network must learn the transport geometry entirely from data, which increases the complexity of the meta-learning task compared to a physically constrained approach.

In practice, the theoretical distinction between PFBR's black-box meta-learning and a physics-constrained approach is expected to manifest in several key performance metrics. First, regarding training sample efficiency, a black-box operator must implicitly learn the underlying transport geometry from vast amounts of synthetic task data. By explicitly embedding the master PDE, a physics-informed model drastically restricts the functional search space, guiding the network directly toward valid transport dynamics. Second, in terms of mode coverage, unconstrained neural operators trained via statistical divergences (e.g., reverse KL divergence) often exhibit mode-seeking behavior, risking mode collapse in multimodal posteriors. A physics-informed flow, conversely, enforces the continuity equation, which acts as a strict conservation of probability mass to prevent the dropping of modes. Finally, while meta-learned operators are prone to overfitting the specific prior and likelihood families seen during training, conditioning the flow on local PDE residuals ensures robust zero-shot generalization to out-of-distribution geometries.

\subsection{Motivation and contribution}

A review of existing methods reveals certain trade-offs in deterministic particle transport for Bayesian inference, which can be categorized into three classes:
\begin{itemize}
    \item Log-homotopy particle flows (e.g., exact \cite{daum2010exact}, incompressible \cite{crouse2021particle}): these methods rely on rigorous derivations based on log-homotopy. However, they are prone to numerical stiffness and instability in strongly nonlinear scenarios, often requiring heuristic regularization techniques for practical implementation.
    \item Interacting particle systems (SVGD \cite{liu2016stein}): this approaches ensures stable convergence, however, comes with high computational costs ($O(N^2)$) due to pairwise particle interactions. Furthermore, operates as online optimization procedure that requires solving a minimization problem at each time step.
    \item Neural operators (PFBR \cite{chen2019particle}): this method offers fast, amortized inference ($O(N)$). However, typically treats the update as a black-box learning task, neglecting the analytical constraints available through the log-homotopy formulation.
\end{itemize}

In this work, we propose a novel framework that integrates log-homotopy particle flow within a deep learning architecture. By interpreting the master PDE as a physical constraint for an amortized neural operator rather than a differential equation to be solved online, we achieve the following contributions:

\begin{itemize}
    \item Physics-informed Bayesian operator: a learning-based operator parameterized by a multilayer perceptron (MLP) is introduced. In contrast to PFBR, which learns an arbitrary flow by minimizing a statistical divergence, this operator is explicitly constrained to satisfy the master PDE. Consequently, optimization is guided by the governing flow dynamics rather than relying solely on data-driven statistical divergences.
    
    \item Unsupervised training scheme: we propose a training objective based on the residual of the master PDE. This formulation enables purely unsupervised training, eliminating the need for ground-truth posterior samples or computationally expensive MCMC simulations.
    
    \item Mitigation of numerical stiffness: we show that the neural parameterization acts as an inherent regularizer, effectively addressing the numerical stiffness that characterizes traditional analytical particle flow filters. The method is evaluated on a highly nonlinear problem featuring a banana-shaped twisted density, where standard numerical integrators for the flow equation typically diverge.
    
    \item Robust mode discovery: we evaluate the proposed method on a four-dimensional Gaussian mixture problem characterized by separated modes. Results indicate that the physics-informed flow prevents mode collapse and identifies all modes compared to baseline methods, while maintaining the computational efficiency of an amortized $O(N)$ update.
\end{itemize}

\section{Log-homotopy particle flow} \label{sec:lhpf}

The fundamental concept of the log-homotopy particle flow (LHPF) filter, introduced by Daum and Huang \cite{daum2007nonlinear, daum2010exact}, is to perform the Bayesian update by transporting particles continuously from the prior distribution $g(x)$ to the posterior distribution $p(x)$ via a flow in pseudo-time.

\subsection{General stochastic formulation}
Let $\lambda \in [0, 1]$ denote a pseudo-time parameter, where $\lambda=0$ corresponds to the prior distribution $g(x)$ and $\lambda=1$ corresponds to the posterior $p(x)$. The evolution of the probability density function $p_{\lambda}(x)$ is defined by the log-homotopy
\begin{equation} \label{eq:homotopy_def}
    \log p_{\lambda}(x) = \log g(x) + \lambda \log h(x) - \log K_\lambda \, ,
\end{equation}
where $h(x)$ is the likelihood function and $K_\lambda = \int g(x) h(x)^\lambda \diff x$ is the normalization constant required to ensure $\int p_{\lambda}(x) \diff x = 1$.

To transform samples from the prior to the posterior, the motion of the particles is modeled by a generic It\^o stochastic differential equation
\begin{equation} \label{eq:general_sde}
    \diff x(\lambda) = f(x, \lambda) \diff \lambda + \sigma(x, \lambda) \diff w_\lambda \, ,
\end{equation}
where $f$ is the drift vector, $\sigma$ is the diffusion matrix, and $w_\lambda$ is a Wiener process. For notational simplicity, the explicit dependence on $x$ and $\lambda$ is omitted when clear from context. The evolution of the probability density $p_\lambda$ under SDE \eqref{eq:general_sde} is governed by the Fokker-Planck equation:
\begin{equation} \label{eq:fpe}
\frac{\partial p_\lambda}{\partial \lambda}
=
-\nabla \cdot (p_\lambda f)
+
\frac{1}{2} \nabla \cdot \left( \nabla \cdot (p_\lambda \sigma \sigma^\top) \right) \, .
\end{equation}

The problem of designing a log-homotopy particle flow is equivalent to finding functions $f$ and $\sigma$ such that the density evolution in \eqref{eq:fpe} matches the boundary conditions imposed by Bayes' theorem and the flow obeys the homotopy constraint \eqref{eq:homotopy_def}. However, since we have one scalar constraint for the evolution of $p_\lambda$, and multiple unknowns as the components of vector $f$ and matrix $\sigma$, the problem is underdetermined. Consequently, there is no unique particle flow; rather, there exists a rich family of valid flows. Various formulations have been derived so far \cite{dai2021new,crouse2020consideration}, including the exact flow with zero diffusion \cite{daum2010exact,toro2023analytic} and stochastic flows \cite{daum2018new}.

\subsection{The deterministic master PDE}
In this work, we consider deterministic flows, where the diffusion matrix $\sigma$ is set to zero. This choice reduces the stochastic evolution \eqref{eq:general_sde} to the ODE
\begin{equation} \label{eq:theODE}
    \frac{\diff x(\lambda)}{\diff \lambda} = f \, .
\end{equation}
Correspondingly, the Fokker-Planck equation simplifies to the continuity equation, describing the conservation of probability mass under a deterministic flow:
\begin{equation} \label{eq:continuity}
    \frac{\partial p_\lambda}{\partial \lambda} = -\nabla \cdot (p_\lambda f) \, .
\end{equation}
To derive the governing equation for the drift $f$, we first differentiate the log-homotopy \eqref{eq:homotopy_def} with respect to $\lambda$:
\begin{equation} \label{eq:homotopy_deriv}
    \frac{\partial \log p_\lambda}{\partial \lambda} = \log h - \frac{\diff \log K_\lambda}{\diff \lambda} \, .
\end{equation}
To obtain the explicit form of the term involving $K_\lambda$, we rely on the constraint that the pdf $p_\lambda$ must remain normalized throughout the flow:
\begin{equation}
    \int p_\lambda \diff x = 1 \, .
\end{equation}
Differentiating both sides with respect to $\lambda$ and assuming regularity conditions that allow the interchange of integration and differentiation, we obtain:
\begin{equation} \label{eq:prob1}
    \frac{\diff }{\diff \lambda} \int p_\lambda \diff x = \int \frac{\partial p_\lambda}{\partial \lambda} \diff x = 0 \, .
\end{equation}
Using the log-derivative identity 
\begin{equation} \label{eq:logderivative}
    \frac{\partial p_\lambda}{\partial \lambda} = p_\lambda \frac{\partial \log p_\lambda}{\partial \lambda} \, ,
\end{equation}
constraint \eqref{eq:prob1} can be expressed as an expectation with respect to the intermediate density $p_\lambda$ as
\begin{equation} \label{eq:expectation_zero}
    \int p_\lambda \frac{\partial \log p_\lambda}{\partial \lambda} \diff x = \mathbb{E}_{p_\lambda} \left[ \frac{\partial \log p_\lambda}{\partial \lambda} \right] = 0 \, .
\end{equation}
This expression simply states that probability mass cannot be created or destroyed throughout the flow.

Substituting the homotopy derivative from \eqref{eq:homotopy_deriv} into \eqref{eq:expectation_zero} yields
\begin{equation} \label{eq:expectation_zero2}
    \mathbb{E}_{p_\lambda} \left[ \log h - \frac{\diff \log K_\lambda}{\diff \lambda} \right] = 0 \, .
\end{equation}
Exploiting the linearity of expectation, and noting that $K_\lambda$ depends only on $\lambda$ we arrive at
\begin{equation} \label{eq:expectation}
    \frac{\diff \log K_\lambda}{\diff \lambda} = \mathbb{E}_{p_\lambda} [\log h] \, .
\end{equation}
This indicates that the rate of change of the log-normalizer is equal to the expected log-likelihood throughout the flow. This result is obtained in \cite{daum2011coulomb} in an alternative way.

By substituting the expectation from \eqref{eq:expectation} back into \eqref{eq:homotopy_deriv} we obtain
\begin{equation} \label{eq:master0}
    \frac{\partial \log p_\lambda}{\partial \lambda} = \log h - \mathbb{E}_{p_\lambda} [\log h] \, .
\end{equation}

Using the log-derivative and the identity $\nabla \cdot (p_\lambda f) = p_\lambda \nabla \cdot f + f \cdot \nabla p_\lambda$ we write the continuity equation \eqref{eq:continuity} in the form
\begin{equation}
     \frac{\partial \log p_\lambda}{\partial \lambda} = - f \cdot \nabla \log p_\lambda - \nabla \cdot f
\end{equation}
and combine it with \eqref{eq:master0} to arrive
at the master PDE of the deterministic log-homotopy flow:
\begin{equation} \label{eq:master_pde}
   -\nabla \cdot f - f \cdot \nabla \log p_\lambda  = \log h - \mathbb{E}_{p_\lambda} [\log h] \, .
\end{equation}
This linear, first-order PDE connects the divergence of the flow field to the deviation of the local log-likelihood from its global mean.

The intermediate density $p_\lambda$ is defined by the log-homotopy \eqref{eq:homotopy_def}. Since $K_\lambda$ does not depend on $x$ we can write
\begin{equation}
    -\nabla \cdot f - f \cdot \nabla (\log g + \lambda \log h )  = \log h - \mathbb{E}_{p_\lambda} [\log h] \, .
\end{equation}

\begin{figure*}[h]
    \centering
    \resizebox{\textwidth}{!}{
    \begin{tikzpicture}[node distance=6.8cm, auto, thick]
        \tikzstyle{block} = [rectangle, draw, fill=blue!5, 
                             text width=4.5cm, text centered, rounded corners, minimum height=2.5cm]
        \tikzstyle{target} = [rectangle, draw, fill=green!10, 
                             text width=4cm, text centered, rounded corners, minimum height=2.5cm, line width=1.5pt]
        \tikzstyle{arrow} = [->, >=stealth, line width=1pt]
        
        \tikzstyle{txt} = [text width=2.5cm, align=center, font=\small\itshape]
        \tikzstyle{lbl} = [align=center, font=\small\itshape] 

        \node [target] (target_pdf) {\textbf{Target density}\\(Boltzmann-Gibbs)\\$p(x) \propto e^{-V(x)}$\\Static distribution};
        
        \node [block, left of=target_pdf] (stoch_pdf) {\textbf{Stochastic evolution}\\(Fokker-Planck equation)\\$ \frac{\partial p_t}{\partial t} = \nabla \cdot (p_t \nabla V) + \Delta p_t$\\Steady state at $t \to \infty$};
        
        \node [block, right of=target_pdf] (homo_pdf) {\textbf{Homotopy evolution}\\(Master PDE)\\ $- \nabla \!\!\cdot \!(p_\lambda f)/p_\lambda \!=\! \log h \!-\! \mathbb{E}_{p_\lambda}[\log h]$ \\ Endpoint at $\lambda = 1$};

        \node [target, below of=target_pdf, node distance=4.5cm] (target_part) {\textbf{Posterior samples}\\(Dirac mixture)\\$\{x_i\}_{i=1}^N \sim p(x)$\\Static samples};

        \node [block, below of=stoch_pdf, node distance=4.5cm] (stoch_part) {\textbf{Brownian motion}\\(Langevin equation)\\$\diff x = -\nabla V \diff t + \sqrt{2} \diff W_t$\\Fluctuating equilibrium};

        \node [block, below of=homo_pdf, node distance=4.5cm] (homo_part) {\textbf{Particle flow}\\(Deterministic ODE)\\$\frac{\diff x}{\diff \lambda} = f(x,\lambda)$\\Characteristic curves};

        \draw [arrow, <->] (stoch_pdf) -- node[txt, left] {Mean-field\\limit} (stoch_part);
        
        \draw [arrow, <->, dashed] (target_pdf) -- node[txt, left] {Empirical\\measure} (target_part);
        
        \draw [arrow, <->] (homo_pdf) -- node[txt, left] {Method of\\characteristics} (homo_part);

        \draw [arrow] (stoch_pdf) -- node[lbl, above] {Relaxation} node[lbl, below] {($t \to \infty$)} (target_pdf);
        
        \draw [arrow, dotted] (stoch_part) -- node[txt, sloped, below] {Ergodicity} (target_part);

        \draw [arrow] (homo_pdf) -- node[lbl, above] {Termination} node[lbl, below] {($\lambda = 1$)} (target_pdf);

        \draw [arrow] (homo_part) -- node[txt, sloped, below] {$\int_0^1$} (target_part);

    \end{tikzpicture}
    }
    \caption{Physics based Bayesian computation. Center: The estimation objective is a static target posterior and its discrete approximation via an empirical measure. Left: Stochastic relaxation, where the system asymptotically converges to a fluctuating equilibrium ($t \to \infty$). The macroscopic and microscopic views are linked via the mean-field limit. Right: Deterministic transport, where the system evolves over a finite horizon ($\lambda \in [0,1]$). The particle trajectories correspond to the characteristic curves of the macroscopic master PDE.}
    \label{fig:inference_landscape}
\end{figure*}
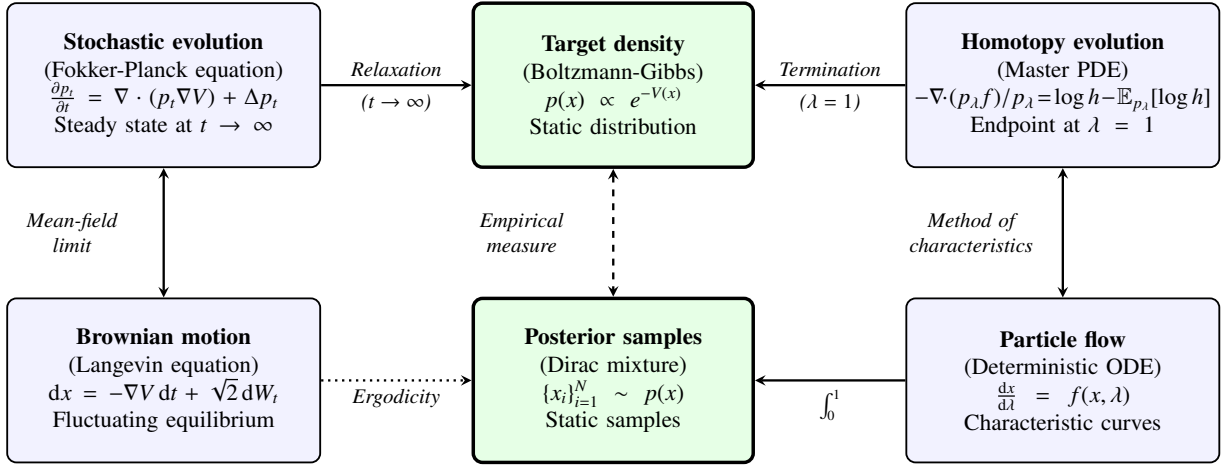

\section{Theoretical analysis} \label{sec:theo}

In this section, we analyze the theoretical properties of the derived master PDE. First, we verify its consistency with the fundamental law of probability mass conservation. Subsequently, we derive the explicit solution for the one-dimensional case, establishing a geometric connection to quantile-preserving transport. Finally, we contextualize the proposed approach within the broader landscape of physics-based Bayesian computation, contrasting the deterministic transport paradigm with stochastic relaxation methods.

\subsection{Consistency with conservation of probability mass}

To verify the physical consistency of the master PDE, we integrate the expression over the entire state space $\mathbb{R}^{n_x}$. First, we rewrite \eqref{eq:master_pde} as
\begin{equation}
    -\frac{1}{p_\lambda} \nabla \cdot (p_\lambda f) = \log h - \mathbb{E}_{p_\lambda}[\log h] \, .
\end{equation}
Multiplying both sides by the density $p_\lambda$ and integrating with respect to $x$, we obtain:
\begin{equation} \label{eq:consistency_integral}
    \int -\nabla \cdot (p_\lambda f) \diff x = \int p_\lambda \left( \log h - \mathbb{E}_{p_\lambda}[\log h] \right) \diff x \, .
\end{equation}
The rhs is 0 due to \eqref{eq:expectation_zero2}. For the lhs, we apply the divergence theorem. Assuming standard regularity conditions where the probability density $p_\lambda$ and the probability current $p_\lambda f$ decay to zero as $\|x\| \to \infty$, the surface integral at the boundary vanishes so we can write
\begin{equation}
    \int \nabla \cdot (p_\lambda f) \diff x = \oint (p_\lambda f) \cdot n \diff S = 0 \, ,
\end{equation}
where $n$ is the unit normal to the surface element $\diff S$.

The above confirms that the master PDE is mathematically consistent with the global conservation of probability mass. It ensures that the net change in probability mass over the entire domain is zero, satisfying the requirement that $\frac{\diff}{\diff \lambda} \int p_\lambda \diff x = 0$.

\subsection{Sanity check in one dimension}

In the one-dimensional case, the master PDE
\begin{equation}
    -\frac{\partial}{\partial x} (p_\lambda f) = p_\lambda \left( \log h - \mathbb{E}_{p_\lambda}[\log h] \right) \,
\end{equation}
has a unique explicit solution. Since we have
\begin{equation}
    p_\lambda \left( \log h - \mathbb{E}_{p_\lambda}[\log h] \right) = p_\lambda \frac{\partial \log p_\lambda}{\partial \lambda} = \frac{\partial p_\lambda}{\partial \lambda} \, ,
\end{equation}
the PDE to be solved is a continuity equation in one dimension:
\begin{equation}
    -\frac{\partial}{\partial x} (p_\lambda f) = \frac{\partial p_\lambda}{\partial \lambda} \, .
\end{equation}
We integrate both sides with respect to the spatial coordinate from $-\infty$ to $x$. Assuming $p_\lambda$ and $p_\lambda f$ vanish at infinity and regularity conditions that allow us to interchange the order of differentiation and integration, we obtain
\begin{equation}
    -p_\lambda f = \frac{\partial}{\partial \lambda} \int_{-\infty}^x p_\lambda \diff u \, .
\end{equation}
The integral on the rhs stands for the cumulative distribution function (cdf) $F_\lambda$ of the intermediate density. The velocity field $f$ is then expressed as
\begin{equation} \label{eq:1d_exact_flow}
    f = -\frac{1}{p_\lambda} \frac{\partial F_\lambda}{\partial \lambda} \, ,
\end{equation}
meaning that in one dimension, the flow is uniquely determined by the homotopy evolution of the pdf and cdf.

The total derivative of $F_\lambda$ is
\begin{equation}
    \frac{\diff F_\lambda}{\diff \lambda} = \frac{\partial F_\lambda}{\partial \lambda} + \frac{\partial F_\lambda}{\partial x}  \frac{\diff x}{\diff \lambda} \, ,
\end{equation}
which reduces to
\begin{equation}
    \frac{\diff F_\lambda}{\diff \lambda} = -p_\lambda f + p_\lambda f = 0 \, .
\end{equation}
Physically, this ensures that the flow preserves the quantile ordering of the particles, which corresponds to the monotone transport.

Similar approaches are derived in \cite{hanebeck2003progressive,schrempf2006dirac} using linear homotopy.

\begin{table*}[ht]
\centering
\caption{Overview of particle-based Bayesian inference methods in dimension $D$ using $N$ particles and $L$ network layers.}
\label{tab:methods_comparison}
\renewcommand{\arraystretch}{1.5}
\footnotesize
\begin{tabular}{@{}p{0.1\textwidth} p{0.13\textwidth} p{0.13\textwidth} p{0.13\textwidth} p{0.13\textwidth} p{0.13\textwidth} p{0.13\textwidth}@{}}
\toprule
\textbf{Feature} 
& \textbf{MCMC} 
& \textbf{LHPF}
& \textbf{SVGD}
& \textbf{NF / CNF} 
& \textbf{PFBR}
& \textbf{PINPF (Ours)} \\
\midrule

\textbf{Work principle} 
& Ergodic theory 
& Log-homotopy 
& Variational inference
& Measure transport
& Meta-learning
& Physics-informed learning \\
\hline

\textbf{Dynamics} 
& Stochastic Markov chain 
& ODE or SDE 
& Deterministic gradient flow 
& Deterministic map / ODE 
& Deterministic neural ODE 
& Deterministic neural ODE \\
\hline

\textbf{Update logic} 
& Accept / reject 
& Numeric integration
& Gradient descent \& kernelized repulsion 
& Apply map / integrate ODE 
& Forward pass of learned $f_\theta$ 
& Forward pass of learned $f_\theta$  \\
\hline

\textbf{Complexity \newline (per step)} 
& $O(N \cdot D^2)$ \newline (likelihood)
& $O(N \cdot D^3)$ \newline (matrix inversion)
& $O(N^2 \cdot D)$ \newline (pairwise kernel)
& $O(N \cdot D^2 \cdot L)$ \newline (network depth)
& $O(N \cdot D^2 \cdot L)$ \newline (network depth)
& $O(N \cdot D^2 \cdot L)$ \newline (network depth) \\
\hline

\textbf{Integration \newline horizon} 
& $t \to \infty$ \newline (asymptotic)
& $\lambda \in [0,1]$ \newline (stiff)
& Iterative opt. \newline (until convergence)
& $1$ step (map) /  $\int_0^1$
& $t \to \infty$ \newline (asymptotic)
& $\lambda \in [0,1]$ \newline (discrete steps) \\
\hline

\textbf{Training} 
& No 
& No 
& No
& Yes
& Yes
& Yes \\
\hline

\textbf{Main \newline advantages} 
& Asymptotically exact 
& Analytic form, no resampling 
& Deterministic, closed-form update 
& Flexible geometry, fast sampling 
& Fast amortized inference 
& Physics-constrained, reduces stiffness \\
\hline

\textbf{Main \newline drawbacks} 
& Correlated samples, slow mixing 
& Numerical stiffness, stability issues 
& High computational cost per step 
& Lacks physical interpretability 
& Black-box, generalization gap 
& Training complexity \\
\bottomrule
\end{tabular}
\end{table*}

\subsection{Overview of particle-based posterior estimation} \label{sec:overview}

Figure~\ref{fig:inference_landscape} shows the scheme of physics based computation of the posterior distribution and categorizes methods based on their representation and their convergence mechanism. Dashed arrow means convergence in distribution, dotted arrow indicates ergodic convergence, and solid arrows stand for analytic convergences.

The objective of Bayesian inference is to characterize the static posterior density $p(x)$, shown in the top center block. The target density can be written in the Boltzmann--Gibbs form $ p(x) \propto  e^{-V(x)}=g(x)h(x)$ where $V(x)$ is the potential function of the distribution. Numerical methods approximate this density using a Dirac mixture formed by a set of discrete samples $\{x_i\}_{i=1}^N$. The samples form an empirical measure that represents the target distribution.

Traditional MCMC and Langevin dynamics (left column) operate via stochastic relaxation. These methods treat the probability density as a physical system relaxing toward equilibrium. Macroscopically, the evolution of the density is governed by the Fokker--Planck equation, which includes a diffusion term. Convergence to the target posterior happens asymptotically as $t \to \infty$. Microscopically, particles follow stochastic differential equations. Even at equilibrium, particles continue to fluctuate because of the Brownian driving force. The connection between the dynamic trajectory and the static samples is established via  ergodicity, where the time-average of a single trajectory converges to the ensemble average of the target density. The vertical correspondence between the SDE and the Fokker--Planck equation is known as the mean-field limit as particle number $N \to \infty$.

The right column shows deterministic particle flow methods based on log-homotopy. These methods view the Bayesian update as a mass transport problem defined by a log-homotopy over a finite pseudo-time interval $\lambda \in [0, 1]$. Macroscopically, the density evolves according to the continuity equation, which enforces the conservation of probability mass without diffusion. Microscopically, particles move according to a deterministic ODE. The flow terminates exactly at $\lambda=1$, mapping the prior ensemble directly to the posterior ensemble. The vertical connection in this paradigm is the method of characteristics, where the particle trajectories are the characteristic curves of the continuity equation.

The framework proposed in this paper explicitly exploits this correspondence: we seek to learn the microscopic particle dynamics such that the evolution of the probability density satisfies the master PDE. By embedding this PDE into the loss function of a neural network, we ensure that the learned particle trajectories constitute a valid deterministic transport map to the target posterior.

Table \ref{tab:methods_comparison} compares the theoretical principles and computational complexity of particle-based methods.
A critical differentiator is the online scaling with respect to the state dimension $D$ and particle number $N$. Analytical LHPF methods typically require the inversion of a covariance matrix or Hessian at every integration step \cite{daum2010exact}, resulting in cubic scaling $O(N D^3)$. In contrast, neural operator-based methods (such as PFBR, NF, and PINPF) rely on feed-forward passes involving matrix-vector multiplications, which has $O(N D^2 L)$ complexity, where $L$ denotes the number of layers (network depth). Because $L$ is typically a small, fixed architectural constant, the effective scaling remains $O(ND^2)$. This quadratic scaling makes neural approaches tractable in high-dimensional settings where analytical flows can be computationally infeasible.

\section{Physics-informed neural particle flow} \label{sec:pinpf}

In this section, we introduce the physics-informed neural particle flow (PINPF). Instead of attempting to solve the stiff differential equations of the particle flow online, our approach approximates the velocity field $f$ using a neural network. This network is trained offline to satisfy the master PDE, resulting in an amortized inference operator that is both computationally efficient and numerically stable.

\subsection{The physics-informed loss function}

We interpret the master PDE \eqref{eq:master_pde} as a physical constraint which expresses a balance between the required evolution of the density and the actual transport induced by the flow field:
\begin{equation} \label{eq:balance_eq}
     \underbrace{-\nabla \cdot f - f \cdot \nabla \log p_\lambda}_{\text{Transport rate}}  = \underbrace{\log h - \mathbb{E}_{p_\lambda} [\log h]}_{\text{Homotopy drive}} \, .
\end{equation}
The rhs of \eqref{eq:balance_eq} originates from the derivative of the log-homotopy equation \eqref{eq:homotopy_deriv} and it drives the flow. The magnitude of this driving force depends on the parameterization of the homotopy. In our formulation, we employ the straight linear homotopy $\lambda$. Had we chosen a general homotopy $\alpha(\lambda)$ (s.t. $\alpha(0) = 0, \alpha(1) = 1$) instead of the linear interpolation, the driving term would be scaled by the derivative $\frac{\diff \alpha}{\diff \lambda}$ as a factor of $\log h$. The effect of the chosen homotopy function on the flow is discussed in \cite{dai2022stiffness}.

The lhs of \eqref{eq:balance_eq} represents the actual change in log-density resulting from the particle motion. It is composed of two physical mechanisms derived from the continuity equation: the compression of the volume element ($-\nabla \cdot f$) and the advection of the density along the flow ($-f \cdot \nabla \log p_\lambda$).

We define the velocity field approximation as a function $f_\theta(\cdot)$, parameterized by a neural network with weights $\theta$.

We define the velocity field approximation as a function $f_\theta$, parameterized by a neural network with weights $\theta$. Notably, because the flow operates in an unbounded state space $\mathbb{R}^{d_x}$, we must ensure that the boundary condition for mass conservation satisfied, that is the probability current $p_\lambda f_\theta$ decays to zero as $\| x \| \rightarrow \infty$. In our framework, this is achieved implicitly without requiring explicit boundary penalty terms in the loss function. For standard exponential-family distributions (e.g., Gaussians), the density $p_\lambda$ decays exponentially toward infinity. Because the MLP utilizes SiLU activations, the predicted velocity $f_\theta$ exhibits at most linear growth in extrapolation regions. Consequently, the exponential decay of the density strictly dominates the network's growth, guaranteeing that $\lim_{\|x\| \rightarrow \infty} p_\lambda(x) f_\theta(x) = 0$ and preventing any probability mass from leaking at the domain boundaries.

The PDE residual $\mathcal{R}(x, \lambda; \theta)$ is the difference between the two sides of \eqref{eq:balance_eq}:
\begin{equation}
    \mathcal{R}(x, \lambda; \theta) = \log h - \mathbb{E}_{p_\lambda}[\log h] - \left( -\nabla \!\cdot\! f_\theta - f_\theta \! \cdot \! \nabla \log p_\lambda \right) .
\end{equation}
The instantaneous loss at $\lambda$ is the expected squared residual over the intermediate density $p_\lambda$:
\begin{equation}
   \mathbb{E}_{p_\lambda}\left[ (\mathcal{R}(x, \lambda; \theta))^2  \right] \, ,
\end{equation}
and throughout the flow the total loss is
\begin{equation}
   \mathcal{L}(\theta) = \int_0^1 \mathbb{E}_{p_\lambda}\left[ (\mathcal{R}(x, \lambda; \theta))^2  \right] \diff \lambda \, .
\end{equation}
In practice, we approximate the expectation using a batch of $N$ particles $\{x_i\}_{i=1}^N$ sampled from the prior and sum up through the $L$ discrete steps of the flow. The empirical loss function is then
\begin{equation} \label{eq:loss_final}
    \mathcal{L}(\theta) = \sum_{k=1}^{L} \frac{1}{N} \sum_{i=1}^N \left( \mathcal{R}(x_i, \lambda_k; \theta) \right)^2 \, .
\end{equation}
Note that the squared residuals are weighted equally across all pseudo-time steps. Empirical testing confirmed that uniform weighting performs best. Down-weighting any segment of the integration horizon allows early or late violations of the continuity equation.

For lower-dimensional state spaces, the exact divergence $\nabla \cdot f_\theta$ is computed via automatic differentiation, which requires $D$ backward passes and scales as $O(ND^3L)$ during training. For scaling the training process to higher dimensions ($D \ge 10$), we substitute the exact divergence with Hutchinson's trace estimator (see Section \ref{sec:high-d-nonlinear}). It is important to note, however, that during online inference, the divergence computation is not required, as the particles are simply propagated through the network's forward pass.

For state dimensions $d_x>1$, Eq. (31) remains underdetermined because it provides a single scalar constraint for the $d_x$ components of the velocity field $f$. According to the Helmholtz decomposition, adding any divergence-free vector field yields another valid transport map. While Daum and Huang \cite{daum2010exact} resolve this ambiguity using explicit structural assumptions (e.g., minimum norm or gradient flow), our approach relies on the implicit regularization of neural networks. Due to their well-documented spectral bias \cite{rahaman2019spectral,zhi2020frequency}, multilayer perceptrons trained via gradient descent naturally favor low-frequency, smooth functions. This bias prevents the network from learning spurious divergence-free motions, allowing it to converge toward a smooth diffeomorphism without requiring rigid analytical constraints.

\subsection{Feature construction}

A critical design choice in learning the velocity field is the selection of the input context for the neural network. While a standard MLP taking only $(x, \lambda)$ as input is theoretically universal, we find that incorporating geometric information significantly improves convergence and generalization.
To determine a necessary feature set, we rely on the theoretical structure of specific Daum--Huang filters, where closed-form solutions for $f$ exist under specific assumptions. We construct our context vector $c_i$ to explicitly include terms appearing in these closed-form solutions. Consequently, the augmented input vector for the $i$-th particle at integration step $\lambda$ is defined as:
The augmented input vector $c_i$ for the $i$-th particle is defined as:
\begin{equation} \label{eq:mlp-features}
    c_i = \left[ x_i, \, \lambda, \, z, \, \log h(x_i), \, \nabla \log p_{\lambda}(x_i), \, \nabla \log h(x_i) \right] \, ,
\end{equation}
where for $\log p_{\lambda}(x)$ we can use the unnormalized intermediate log-density $\log g(x) + \lambda \log h(x)$.
The components are chosen with specific roles: $x_i$, $\lambda$, and $z$ establish the scale of the state and the measurement space, while $\log h$ provides direct likelihood feedback. Crucially, we include the gradients of the likelihood and the intermediate potential. By explicitly providing these gradients, the network $f_\theta$ learns the geometry of the problem, rather than simply overfitting the state space.

\subsection{Training} \label{sec:training}

Unlike Neural ODEs, which typically backpropagate gradients through the entire integration trajectory using the adjoint method, we treat the training of the flow at each discretization step $\lambda_k$ as an independent regression problem. In standard continuous normalizing flows, backpropagation through time is necessary because the loss is evaluated only at the terminal time ($\lambda=1$). By contrast, our loss function is the master PDE, which acts as an instantaneous constraint. Because the global log-homotopy trajectory is completely determined by satisfying this local continuity equation everywhere in pseudo-time, the network does not need to perform long-range planning.

The particles are propagated numerically from $\lambda=0$ to $\lambda=1$ using the Euler method:
\begin{equation}\label{eq:euler}
    x_{k+1} = x_k + f_\theta(c_k) \Delta \lambda.
\end{equation}
During training, a fixed pseudo-time step $\Delta \lambda$ is used to provide a stable optimization target. At each step $k$, the loss $\mathcal{L}(\theta)$ is computed based on the current particle positions $x_k$, and gradients are applied to update $\theta$. The computational graph is detached after the update step. This prevents the accumulation of gradients across time steps, significantly reducing memory consumption and avoiding the vanishing/exploding gradient problem common in training deep dynamical systems. This approach effectively breaks the global transport problem into a sequence of local physics-informed constraints without deteriorating the global validity of the transport map.

The training procedure is detailed in Algorithm \ref{alg:neural-flow-training}. In each iteration, we sample a batch of inference tasks from a training dataset $\mathcal{D}_{\mathrm{train}}$. Each task is defined by a measurement realization $z$ and a parameter set governing the $g(x)$ prior and the $h(x)$ likelihood functions. This multi-task framework, inspired by \cite{chen2019particle}, enables the model to learn a generalized Bayesian update operator rather than solving a single instance. For every task, a particle set $\{x_i\}_{i=1}^N$ is initialized by sampling the prior $g(x)$.

\subsection{Inference dynamics}

During the evaluation/inference phase, we adopt the adaptive step size selection scheme
in the Euler method \eqref{eq:euler} from Mori et al. \cite{mori2016adaptive}. To mitigate numerical stiffness, the step size is adjusted such that the particle displacement in state space is bounded by a constant threshold $\Delta L$:
\begin{equation}
    \Delta \lambda = \frac{\Delta L}{\max_i \|f_\theta(c_k)\|}.
\end{equation}
This adaptive scheme directly determines the number of function evaluations (NFE) during the integration of the flow, which is the primary factor influencing computational time. While analytical Daum--Huang flows often exhibit extreme stiffness requiring a very high NFE (i.e. generally small $\Delta\lambda$) to maintain stability, our learned neural flow tends to be smoother, allowing for a significantly lower NFE and faster inference.
The adaptive $\Delta \lambda$ scheme is only used during inference. Applying it during training would couple the integration grid to the currently unstable velocity field, restricting convergence.

\subsubsection{Design flexibility}
While our implementation uses absolute state coordinates and independent particle processing, the neural framework is flexible enough to incorporate more complex input structures. For instance, to enforce scale-invariance, one could employ nondimensionalization techniques or formulate the problem as an error-state estimation. Furthermore, the choice of coordinate system can play a significant role in flow stability. As noted by Crouse \cite{crouse2021particle}, standard Cartesian coordinates can introduce artificial singularities or biases in specific applications (e.g., bearings-only tracking), whereas transforming the state into polar coordinates can simplify the flow. Finally, while our MLP treats particles independently, the architecture could be extended using deep sets or attention mechanisms to condition on population-level features in highly non-Gaussian regimes.

\begin{algorithm}[h!]
\caption{Physics-Informed Neural Particle Flow Training}
\label{alg:neural-flow-training}
\begin{algorithmic}[h]
\REQUIRE Dataset of tasks $\mathcal{D}$, number of particles $N$, pseudo step size $\Delta \lambda$
\STATE Initialize MLP parameters $\theta$
\WHILE{training}
    \STATE \textbf{Task Sampling:}
    \STATE Sample batch of tasks $\{z, g(x), h(z|x)\} \sim \mathcal{D}_{\mathrm{train}}$
    \STATE Sample initial particles $x_{0}^{i} \sim g(x)$ for $i=1, \ldots, N$
    \STATE Initialize total loss $\mathcal{L}_{\text{total}} \leftarrow 0$
    \STATE
    \STATE \textbf{Flow Integration:}
    \FOR{$k = 0$ \textbf{to} $1/\Delta \lambda - 1$}
        \STATE $\lambda_k \leftarrow k \cdot \Delta \lambda$
        \FOR{each particle $x_k^{i}$}
            \STATE \textit{// Compute log-homotopy potential}
            \STATE $\log p(x_k) \leftarrow \log g(x_k) + \lambda_k \log h(z|x_k)$
            \STATE
            \STATE \textit{// Feature construction \eqref{eq:mlp-features}}
            \STATE Compute gradients via autodiff:
            \STATE $v_p = \nabla \log p(x_k)$, $v_h = \nabla \log h(z|x_k)$
            \STATE Construct input $c_k = [x_k, \lambda_k, z, \log h, v_p, v_h]$
            \STATE
            \STATE \textit{// Predict flow vector}
            \STATE $f_k \leftarrow \text{MLP}_\theta(c_k)$
        \ENDFOR
        \STATE
        \STATE \textit{// Compute physics-informed loss}
        \STATE Compute divergence $\nabla \cdot f_k$ via autodiff
        \STATE Estimate expectation $\mathbb{E}[\log h] \approx \frac{1}{N} \sum_{i} \log h(z|x_k^i)$
        \STATE $\mathcal{L}_1 \leftarrow \log h(z|x_k) - \mathbb{E}[\log h]$
        \STATE $\mathcal{L}_2 \leftarrow - \nabla \cdot f_k - (\nabla \log p(x_k)) \cdot f_k$
        \STATE $\mathcal{L}_{\text{step}} \leftarrow \| \mathcal{L}_1 - \mathcal{L}_2 \|^2$
        \STATE $\mathcal{L}_{\text{total}} \leftarrow \mathcal{L}_{\text{total}} + \mathcal{L}_{\text{step}}$
        \STATE
        \STATE \textit{// Particle update (Euler step)}
        \STATE $x_{k+1} \leftarrow x_k + f_k \cdot \Delta \lambda$
        \STATE Detach $x_{k+1}$ from computation graph
    \ENDFOR
    \STATE
    \STATE \textbf{Optimization:}
    \STATE Update $\theta \leftarrow \text{Adam}(\nabla_\theta \mathcal{L}_{\text{total}})$
\ENDWHILE
\end{algorithmic}
\end{algorithm}

\section{Experiments} \label{sec:exp}

\subsection{Evaluation metrics}

To quantitatively assess the discrepancy between the estimated particle ensemble and the true posterior, we employ two complementary metrics: the energy distance (ED) \cite{szekely2013energy} and the sliced Wasserstein distance (SWD) \cite{bonneel2015sliced}. We selected these because they are distribution-free and parameter-free metrics that can be applied directly on samples without tuning kernel or bandwidth parameters.

Let $X=\left\{x_i\right\}_{i=1}^N$ denote the set of particles generated by an inference algorithm, and $Y=\left\{y_j\right\}_{j=1}^M$ denote a set of ground-truth samples drawn from the reference posterior. 

The energy distance provides a robust, global measure of statistical equality between multivariate distributions. The squared empirical ED is computed as:
\begin{multline}
    \text{ED}^2(X, Y) = \frac{2}{NM} \sum_{i=1}^N \sum_{j=1}^M \|x_i - y_j\|_2  \\
    - \frac{1}{N^2} \sum_{i=1}^N \sum_{j=1}^N \|x_i - x_j\|_2 - \frac{1}{M^2} \sum_{i=1}^M \sum_{j=1}^M \|y_i - y_j\|_2 \, .
\end{multline}
In practice we set $M=N$ by sampling the ground truth $N$ times.

Complementing this, the sliced Wasserstein distance captures structural similarities in high-dimensional spaces. Standard Wasserstein distances scale poorly with dimension ($O(N^3)$), but the 1D Wasserstein distance can be computed in $O(N \log N)$ time by simply sorting the samples. The SWD projects the $d$-dimensional samples onto 1D lines and averages the 1D distances over random directions on the unit sphere $\mathbb{S}^{d-1}$. 
Let $\{e_l\}_{l=1}^L$ be a set of $L$ random unit vectors sampled uniformly from $\mathbb{S}^{d-1}$. We project the particles onto each vector $e_l$ and sort them such that $x^{e_l}_{1} \le \dots \le x^{e_l}_{N}$ and $y^{e_l}_{1} \le \dots \le y^{e_l}_{N}$. The squared empirical SWD of order 2 is approximated as:
\begin{equation}
    \text{SWD}^2(X, Y) \approx \frac{1}{L} \sum_{l=1}^L \left( \frac{1}{N} \sum_{i=1}^N \left| x^{e_l}_{i} - y^{e_l}_{i} \right|^2 \right) \, .
\end{equation}
Together, these metrics yield a comprehensive evaluation of both the local accuracy and the global mode coverage achieved by the inference algorithms.

\subsection{Four-dimensional Gaussian mixture posterior}\label{sec:4d-gauss}

\begin{table}[ht]
    \centering
        \caption{Hyperparameter settings for training and inference of PINPF.}
    \label{tab:neural_params}
    \begin{tabular}{@{}lr@{}}
        \toprule
        \textbf{Parameter} & \textbf{Value} \\
        \midrule
        \multicolumn{2}{l}{\textit{MLP Architecture}} \\
        Hidden Layers & 6 \\
        Hidden Dimension & 64 \\
        Activation Function (each layer) & SiLU \\
        
        \midrule
        \multicolumn{2}{l}{\textit{Training}} \\
        Num. particles & 500 \\
        Pseudo-time step & fix, $\Delta\lambda = 0.01$ \\
        Optimizer & Adam \\
        Training Epochs & 6000 \\
        Learning Rate & 0.004 \\
        LR Decay Rate ($\gamma$) & 0.8 \\
        LR Decay Frequency & 300 epochs \\
        Gradient Clipping & 1.0 \\
        Training samples & 1000 \\
        Batch size & 64 \\
        \midrule
        \multicolumn{2}{l}{\textit{Inference}} \\
        Num. particles & 1500 \\
        Pseudo-time step & adaptive, $\Delta L = 0.5$ \\
        \bottomrule
    \end{tabular}
\end{table}

\begin{table}[t]
    \centering
    \caption{Performance results on the four-dimensional Gaussian mixture posterior problem. The mean values are reported over the 100 samples from the test dataset.}
    \label{tab:4dgmm-general}
        \begin{tabular}{@{}lccc@{}}
            \toprule
            Method & ED & SWD & GPU Time [s] \\
            \midrule
            PINPF (ours)          & 0.1150 & 0.4201 & 0.0442 \\
            \midrule
            annealed MCMC & 0.0958 & 0.3854 & 3.8336 \\
            NSF             & 0.1416 & 0.4186 & 0.0499 \\
            SVGD          & 0.1997 & 0.5262 & 1.4170 \\
            Incompressible Flow & 0.2338 & 0.6052 & 0.1257 \\
            \bottomrule
        \end{tabular}%
\end{table}

We evaluate PINPF on a multimodal posterior estimation task involving a four-dimensional state space. The training dataset consists of 1000 inference tasks. For each task, the prior is a zero-centered Gaussian with diagonal covariance, where the diagonal variances are sampled uniformly from $[1, 10]$ for each dimension. The likelihood is defined as a three-component Gaussian mixture model (GMM) with equal component weights. The means of the mixture components are sampled uniformly from $[-3, 3]$ for each dimension, and their diagonal covariances consist of variances sampled uniformly between $0.3^2$ and $0.7^2$. This configuration yields a complex, multimodal Gaussian posterior characterized by occasionally separated modes and varying component importance.

The test set is composed of 100 tasks generated using the same parameter distributions.
Random examples from this test set are shown in Fig. \ref{fig:gmm-matrix}.
We present the quantitative results in Table~\ref{tab:4dgmm-general}.
The training and inference hyperparameters for PINPF, determined via ablation on the validation set, are listed in Table~\ref{tab:neural_params}.

For the evaluation, the ED was computed using $10^4$ ground-truth samples drawn from the analytic posterior. For the SWD, we used 1000 projections and downsampled the analytic ground truth so that its number of particles matched that of the inference methods ($N=1500$).

\paragraph{Baselines}
Since traditional MCMC algorithms struggle to sample from multimodal distributions, we employ an annealed realization. The annealed MCMC method uses a linear annealing schedule $\beta \in [0,1]$ divided into 10 steps. A population of 1500 particles is initialized from the prior. At each temperature step, particle states are updated using 5 sequential iterations of the No-U-Turn Sampler (NUTS)~\cite{hoffman2014no} with a fixed step size of 0.1. 

We also compare against SVGD~\cite{liu2016stein}, which evolves a population of 1500 particles using the Adagrad optimizer. The optimization runs for 500 iterations with a learning rate of 0.2, using a radial basis function kernel with bandwidth determined by the median heuristic.

As an amortized baseline, we trained a conditional neural spline flow (NSF)~\cite{durkan2019neural} on the same dataset that was used for training PINPF.
The flow architecture consists of a stack of 20 coupling layers combined with random permutations to ensure dimensional mixing. To effectively capture the disconnected modes of the target posterior, we use rational quadratic spline transforms with resolution of 32 bins. The spline parameters are predicted by a residual network with two linear layers of 128 hidden units each, which takes the flattened observation, prior parameters, and likelihood parameters as context.

Additionally, the Daum--Huang incompressible flow (IF) is evaluated
with the same number of particles and integration scheme as the PINPF.
Notably, its computational time is 3$\times$ of PINPF's due to the large
number of NFE caused by the stiffness of the incompressible flow.

While the PFBR represents a conceptually related neural operator, it is heavily tailored for sequential dynamic estimation and relies on legacy framework dependencies that limit a stable reproduction in our static, one-step benchmark setup. Therefore, we focus our empirical comparisons on SVGD, annealed MCMC, and the continuous normalizing flow approaches, relying on the extensive theoretical comparison with PFBR provided in Section \ref{sec:overview} and \ref{sec:finiteVSasymp}.

\begin{figure}
    \centering
    \includegraphics[width=1.0\linewidth]{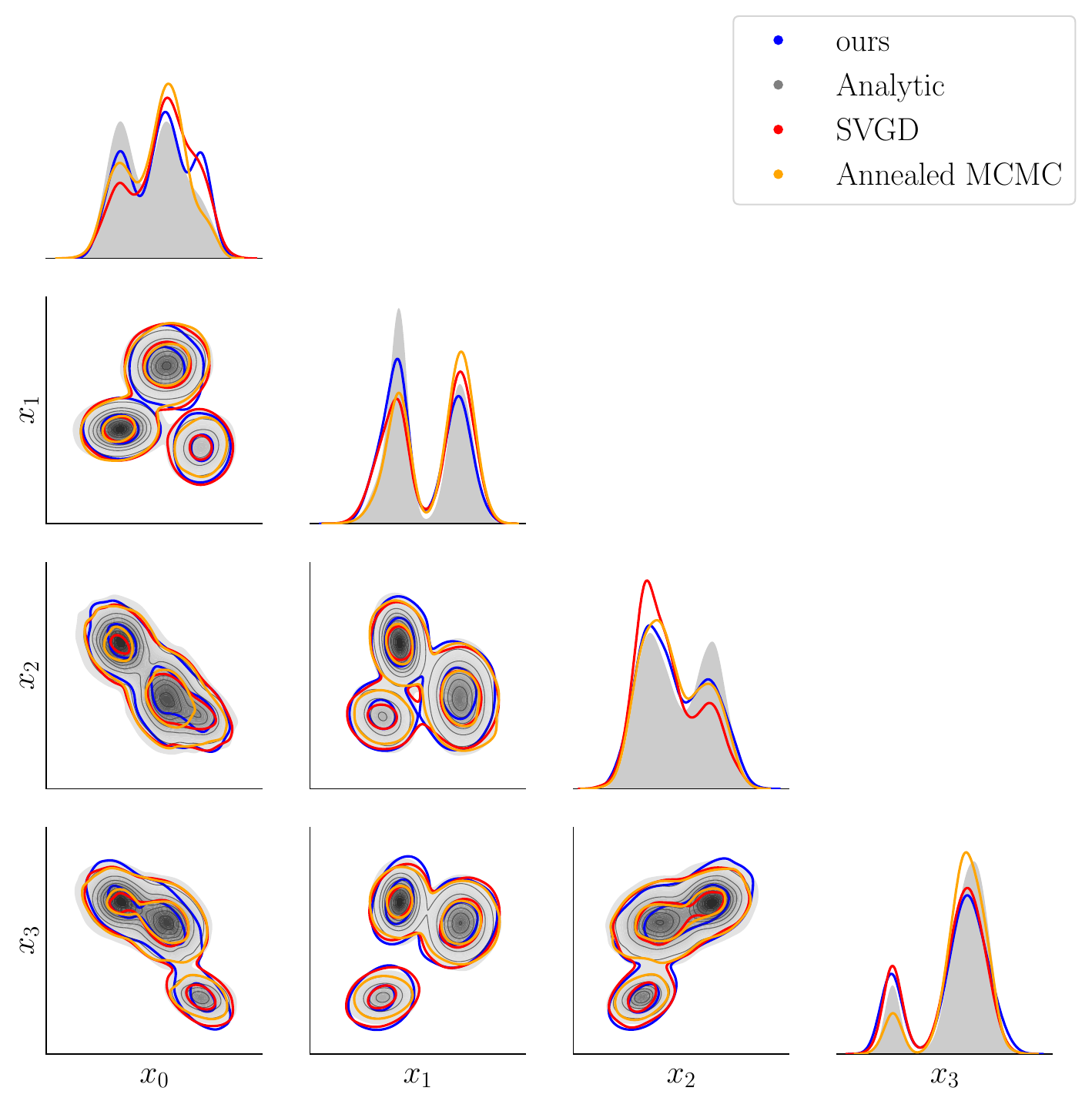}
    \caption{Corner plot for a representative test inference task,
which is selected as the one whose quantitative performance is closest to the mean across all 100 test problems (Table~\ref{tab:4dgmm-general}). Posterior samples from our method, SVGD, and annealed MCMC are shown alongside the analytic reference. Each method generates $1500$ particles, and marginal and pairwise densities are estimated via kernel density estimation.}
\label{fig:4d-qualitative}
\end{figure}

\paragraph{Qualitative analysis}
We present a corner plot in Fig.~\ref{fig:4d-qualitative} to examine mode coverage. To keep the figure uncluttered, we omit the results for IF and NSF, focusing on the comparison between our proposed method and the primary sampling baselines.
Additional qualitative examples are provided in Fig.~\ref{fig:gmm-matrix}.

\begin{figure}
    \centering
    \includegraphics[width=1.0\linewidth]{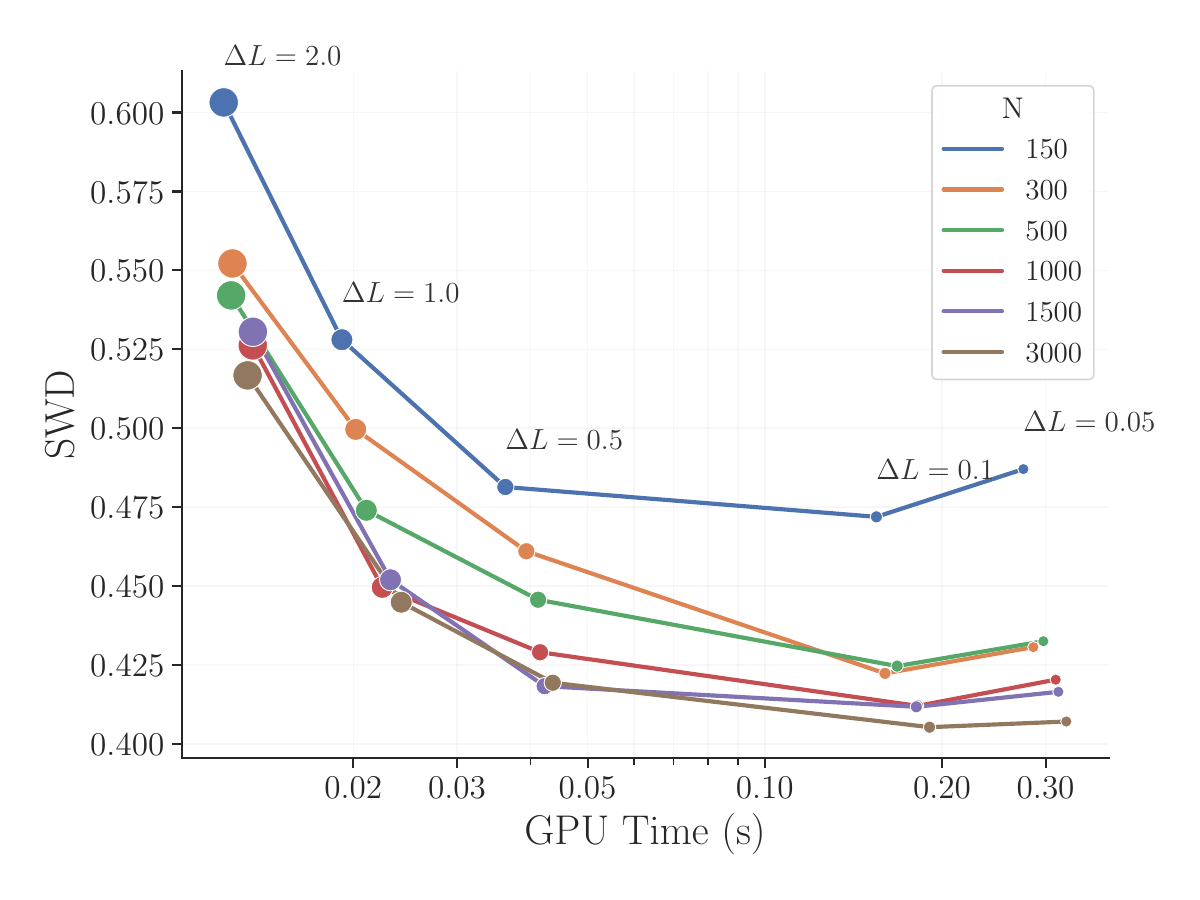}
    \caption{The effect of different adaptive step thresholds $\Delta L$ and particle numbers $N$ on the validation dataset in terms of SWD and computational time.  }
    \label{fig:4d-ablation} 
\end{figure}

\paragraph{Ablation on flow parameters}
We analyze the sensitivity of PINPF to the adaptive step threshold $\Delta L$ and the number of particles $N$ during evaluation. Figure~\ref{fig:4d-ablation}. illustrates the trade-off between GPU runtime and SWD on the validation set. The distinct elbow point observed in the performance curves supports the selection of $N = 1500$ and $\Delta L = 0.5$ for the final evaluation. Notably, the computational cost of increasing $N$ is marginal due to the highly parallelized nature of the flow operations.

\paragraph{Ablation on flow input features}
To validate the feature selection strategy proposed in our methodology (see Eq.~\ref{eq:mlp-features}), we performed an ablation study by training a variant of PINPF where the gradient terms $\nabla \log p$ and $\nabla \log h$ were excluded from the context. 
We observed that omitting gradients results in a significantly slower decline of the loss. Quantitatively, the model using gradient inputs achieved a superior ED of $0.1138$ compared to $0.1274$ for the gradient-free variant. This corresponds to a $10.6\%$ improvement in accuracy. While computing these input derivatives via automatic differentiation introduces a computational overhead (increasing the training time by $15.6\%$) the substantial gain in accuracy and convergence speed justifies this design choice.

\subsubsection{Out-of-distribution prior}

To further assess the robustness and zero-shot generalization capability of the learned flow, we construct an out-of-distribution (OOD) test dataset that mirrors the previous experimental setup, except that the prior distribution is replaced by a GMM with three components. Importantly, PINPF was not retrained or fine-tuned for this experiment.

The GMM prior consists of 3 modes with randomly sampled mixture weights, normalized per instance. The component means are generated by perturbing a common central location with Gaussian noise of standard deviation $2.0$ in each dimension, resulting in clustered but distinct modes. Each component is assigned a diagonal covariance, with variances independently sampled from a uniform distribution on $[1,5]$. As a result, the prior components are deliberately chosen to be broad, so as not to introduce additional well-separated posterior modes, but rather to modify the geometry and gradients of the prior density.

During training, PINPF is conditioned on the prior only through the gradient term $\nabla \left( \log g(x) + \lambda \log h(x) \right)$. This design encourages robustness with respect to moderate changes in the prior, as the network is exposed to prior information exclusively via its score. In the present OOD setting, inference is performed using gradients computed from the GMM prior described above, while the network parameters correspond to training with a unimodal Gaussian prior.

Due to the increased difficulty of the inference problem, we refine the inference parameters by generally reducing the step size via setting $\Delta L = 0.1$, and increase the number of particles to $N = 3000$ for both PINPF and the IF. This choice remains computationally feasible, as PINPF is still substantially faster than the competing baselines. The hyperparameters of annealed MCMC and SVGD are kept unchanged.
NFS would have required retraining because its context depended on the model family used in the prior work; therefore, we excluded it from this study.

The quantitative results are summarized in Table~\ref{tab:ood_results}. As expected, all methods exhibit a degradation in performance under the OOD prior. Nevertheless, PINPF continues to outperform SVGD in both accuracy metrics, while maintaining a significantly lower computational cost. This result highlights the robustness of the learned neural flow to prior misspecification.

\begin{table}[h]
    \centering
    \caption{Zero-shot generalization performance on the out-of-distribution (GMM prior) dataset. PINPF was not retrained.}
    \label{tab:ood_results}
    \begin{tabular}{@{}lccc@{}}
        \toprule
        Method & ED & SWD  & Time [s] \\
        \midrule
        PINPF (ours) & 0.4352 & 0.7045 & 0.2243 \\
        \midrule
        annealed MCMC & 0.1427 & 0.4142 & 4.3955 \\
        SVGD & 0.4521 & 0.7701 & 2.2110 \\
        Incompressible flow & 0.4819 & 0.7329 & 0.3003 \\
        \bottomrule
    \end{tabular}
\end{table}

While PINPF demonstrates robust zero-shot generalization compared to other deterministic baseline flows, a noticeable performance gap remains when compared to Annealed MCMC under the OOD prior (ED: 0.4352 vs 0.1427). This gap highlights a fundamental trade-off in amortized inference. Because MCMC algorithms perform computationally intensive stochastic exploration from scratch for every specific inference task, their performance is largely unaffected by whether a problem is in-distribution or not. In contrast, an amortized neural operator relies on the transport strategies it learned during training. A significant portion of this performance gap could be closed by training the network on a more diverse family of prior distributions (e.g., explicitly including multi-modal or skewed densities during the offline phase). Nevertheless, a fundamental theoretical limitation remains: an amortized, finite-time transport map will naturally exhibit slightly higher approximation errors on radically novel geometries than an instance-specific, asymptotic sampler. This trade-off, however, is heavily justified in practical settings by the roughly 20-fold reduction in computational inference time (0.22\,s vs. 4.40\,s).

\subsection{Single time-difference-of-arrival fusion}

We also evaluate PINPF on a challenging nonlinear tracking scenario \cite{aulia2025navigation,csuzdi2025exact} which involves fusing a single time-difference-of-arrival (TDOA) measurement. As detailed in \cite{mori2016adaptive, crouse2021particle}, this measurement model induces a non-Gaussian, hyperbolic (banana-shaped) likelihood function. When the likelihood is fused with a Gaussian prior, the resulting posterior is highly non-Gaussian, typically exhibiting significant curvature. Furthermore, depending on the location and variance of the prior, the posterior can become multimodal, with separated modes.

\begin{figure}
    \centering
    \includegraphics[width=1.0\linewidth]{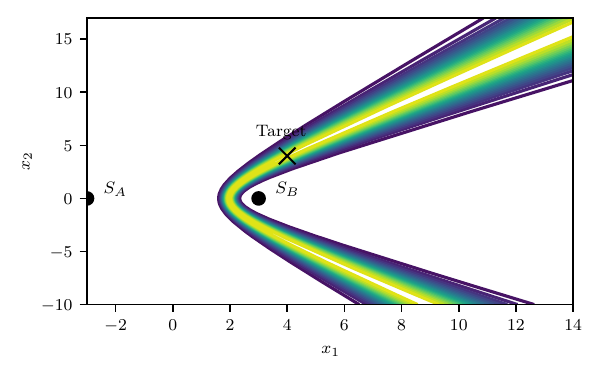}
    \caption{The likelihood landscape of the TDOA measurement. The nonlinear measurement equation creates a hyperbolic high-probability ridge.}
    \label{fig:lik-contour}
\end{figure}

The TDOA measurement $z$ is defined as the difference in Euclidean distances between a target $x \in \mathbb{R}^2$ and two sensors $S_A, S_B \in \mathbb{R}^2$:
\begin{align}
    z &= h(x) + \nu, \\
    h(x) &= \|x-S_A\| - \|x-S_B\|, \label{eq:tdoa}
\end{align}
where the measurement noise is Gaussian $\nu\sim\mathcal{N}(0,\sigma^2)$.
Figure~\ref{fig:lik-contour} shows the likelihood contour for the experimental setup described in \cite{mori2016adaptive}: sensors are located at $S_A=[-3,0]$ and $S_B=[3,0]$, the target is positioned at $[4,4]$ and observed with a noise standard deviation of $\sigma = 0.3$.

\subsubsection{Experimental setup}
We generated a synthetic dataset consisting of 1000 training tasks for PINPF, 100 validation tasks for ablation study, and 100 test tasks for evaluation. For each task, the true target state is sampled from $\mathcal{N}([4, 4]^\top, \mathrm{diag}(1.5^2, 1.5^2))$. The measurement noise standard deviation $\sigma$ is sampled uniformly from $\mathcal{U}[0.4, 0.9]$. To simulate challenging initialization conditions, the prior means are significantly offset from the true state by perturbations drawn from $\mathcal{N}(0, \mathrm{diag}(4^2, 5^2))$, while the diagonal prior variances are sampled from $\mathcal{N}(5, 1)$, rejecting any negative samples. Reference posterior samples for evaluation were generated using grid sampling.
The location of the sensors $S_A,S_B$ are unchanged.

We compare the PINPF against three analytic Daum--Huang flow baselines (incompressible flow, mean exact flow, localized exact flow \cite{crouse2020consideration}), the SVGD, and the annealed MCMC method, which is expected to perform well in this low-dimensional, generally unimodal problem.

Based on the same ablation study described in Section \ref{sec:4d-gauss}., we selected a particle count of $N=1000$ and an adaptive step threshold of $\Delta L = 1.0$ for the flows.

\subsubsection{Results and discussion}

\paragraph{Quantitative analysis}
The performance metrics on the test set are summarized in Table~\ref{tab:performance_summary}. The PINPF significantly outperforms the analytic flow baselines in terms of distribution matching, measured by ED and SWD. 
While the mean exact flow and localized exact flow are computationally inexpensive, they rely on Gaussian approximations that fail to capture the curvature of the TDOA posterior. The incompressible flow, while flexible, is computationally heavier and exhibits higher error. 
The increase in computation is the result of the adaptive stepsize selection algorithm,
which generally assigns small $\lambda$ stepsizes, thus increasing the required 
function evaluations by approximately an order of maginude in this case.
As a result, it can be said, that the learned flow is less stiff,
therefore can use larger stepsizes while still maintaining good performance.
Among the deterministic methods evaluated, PINPF has accuracy closest to the annealed MCMC benchmark and achieves competitive inference time.

\begin{table}[t]
    \centering
    \caption{Performance summary on the TDOA task (averaged over 100 test samples).}
    \label{tab:performance_summary}
        \begin{tabular}{@{}lccc@{}}
            \toprule
            Method & ED  & SWD  & Time [s] \\
            \midrule
            PINPF (ours)               & 0.0697 & 0.3238 & 0.0196 \\
            \midrule
            Incompressible flow       & 0.2415 & 1.0460 & 0.3767 \\
            Local Gaussian exact flow & 0.1749 & 0.6155 & 0.0132 \\
            Mean Gaussian exact flow  & 0.2521 & 0.6277 & 0.0101 \\
            SVGD                      & 0.1344 & 0.4084 & 0.3733 \\
            annealed MCMC             & 0.0036 & 0.1527 & 4.8788 \\
            \bottomrule
        \end{tabular}%
    \end{table}

\paragraph{Qualitative analysis}
We analyze two distinct scenarios to understand the failure modes of the Daum--Huang flow baselines.

\begin{figure}
    \centering
    \includegraphics[width=1.0\linewidth]{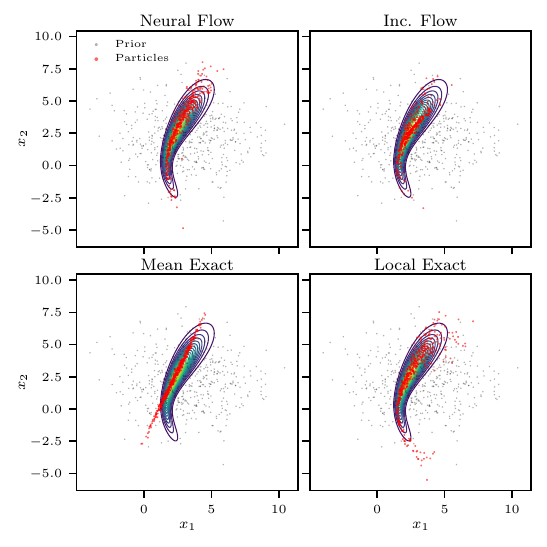}
    \caption{Qualitative comparison on a sample with an informative prior. The analytic flows struggle to capture the ``banana'' shape (Mean Exact) or properly cover the variance (Incompressible), while the Neural Flow matches the posterior geometry accurately.}
    \label{fig:tdoa-general}
\end{figure}

\begin{figure}
    \centering
    \includegraphics[width=1.0\linewidth]{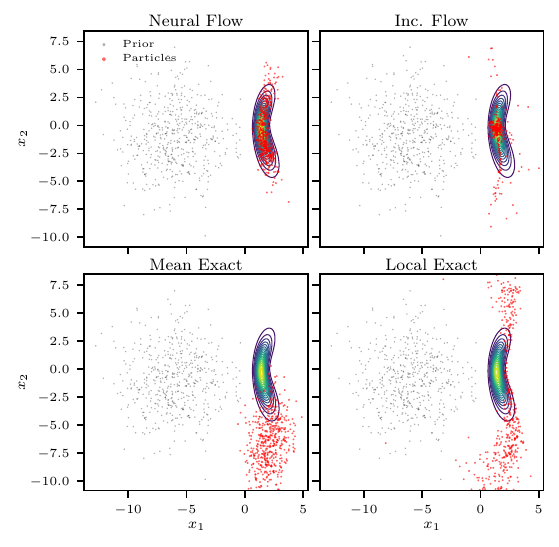}
    \caption{Comparison on a sample with a distant prior. The Gaussian-based approximations (Exact flows) fail due to linearization errors. The Neural Flow robustly transports the particles to the posterior.}
    \label{fig:tdoa-wrong-prior}
\end{figure}

\begin{itemize}
    \item {Scenario A: Informative prior (Fig.~\ref{fig:tdoa-general}).} 
In this case, the prior is reasonably close to the likelihood. 
As expected, the mean exact flow produces samples from a Gaussian distribution, 
which fails to capture the tails. 
The incompressible flow demonstrates the ``fringe'' behavior described by Fig.~8 in \cite{mori2016adaptive}. Particles move towards the likelihood but stop at the fringe of the probability mass, meaning they accumulate at the outer boundary of the high-probability region rather than penetrating the mode, which leads to poor variance estimation and an artificial ring-like particle distribution.
The localized exact flow performs better by linearizing locally, but it overestimates the density in the lower tail region.
This local linearization, however only performs well in this case,
as the prior is informative.
In contrast, PINPF correctly moves the particles into the non-Gaussian curved shape, though with slight variance underestimation near the mode.
\item {Scenario B: Distant prior (Fig.~\ref{fig:tdoa-wrong-prior}).}
This scenario represents a difficult test where the prior mean is far from the true target. 
The flows based on linearization (mean exact flow and localized exact flow) collapse, as the linearization of the measurement function produces inaccurate results when the prior
particles are in very low likelihood regions. This result aligns with Fig. 3/a in \cite{crouse2021particle}.
The incompressible flow concentrates on the mode, but also scatters particles in incorrect directions.
PINPF demonstrates better posterior distribution coverage. Because the network approximates the global transport map via amortization, it successfully navigates particles from the distant prior to the correct posterior mode without the linearization errors occurring in the exact flows.
\end{itemize}

\begin{figure}
    \centering
    \includegraphics[width=1.0\linewidth]{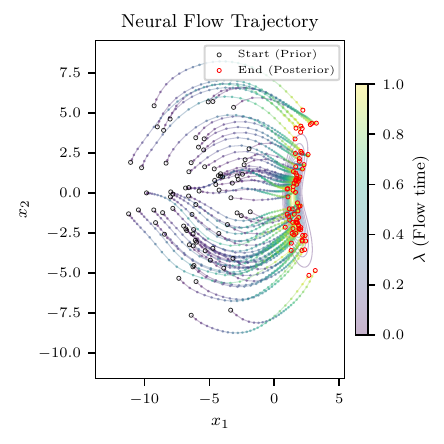}
    \caption{Particle trajectories generated by PINPF for the sample in Fig.~\ref{fig:tdoa-wrong-prior}. The color gradient represents the flow time $\lambda \in [0, 1]$. The model learns a smooth transport map that compresses the variance along the informative direction of the measurement.}
    \label{fig:tdoa-trajectory}
\end{figure}

\paragraph{Trajectory analysis}
Figure~\ref{fig:tdoa-trajectory} illustrates the particle trajectories generated by PINPF. The learned vector field produces smooth paths that particles follow from the prior to the posterior.  Crucially, because any local violation of the continuity equation would compound during integration and distort the final distribution, the accurate formation of the complex posterior geometry implicitly confirms that the network maintains a stable, low PDE residual even when traversing regions far from the prior mean.

\subsubsection{Generalization to dynamic sensor geometries}
The current TDOA experiment utilizes a fixed sensor configuration to evaluate the fundamental geometry of the filtering problem. However, in real-world tracking, sensor positions often vary. Changing the sensor geometry effectively alters the eccentricity of the hyperbolic likelihood and applies a rigid translation and rotation to the measurement landscape. To generalize PINPF to such dynamic settings, the network context vector $c_i$ can be augmented with the sensor coordinates $S_A$ and $S_B$. Alternatively, the rotational degrees of freedom can be eliminated entirely by projecting the particle states into a local, sensor-aligned coordinate frame prior to the flow integration. Furthermore, because our proposed feature vector explicitly evaluates the local likelihood gradient $\nabla \log h$, the model inherently captures the localized curvature and relative orientation of the measurement geometry, making it uniquely suited to handle shifting sensor configurations.

\subsection{High-dimensional nonlinear problem}\label{sec:high-d-nonlinear}

To address scalability regarding the dimensionality of the state space and measurement space, we evaluate PINPF on a high-dimensional nonlinear inference problem with state and measurement dimensions $D = 10$ and $D = 15$. This experiment demonstrates that the proposed method generalizes beyond the low-dimensional settings considered so far, while retaining its computational efficiency and accuracy advantages.

For these higher-dimensional settings, computing the exact divergence $\nabla \cdot f_\theta$ in the master PDE residual requires $D$ separate backward passes, scaling training complexity to $O(ND^3L)$. To mitigate this, we employ Hutchinson's trace estimator~\cite{hu2024hutchinson}, which approximates the divergence as $\nabla~\cdot~f_\theta~\approx~v^\top (\partial f_\theta / \partial x)\, v$, where $v \in \{-1, +1\}^D$ is a Rademacher random probe vector. This reduces the estimate to a single vector-Jacobian product, restoring $O(ND^2L)$ training complexity. The estimator is unbiased in expectation. Note that during online inference the divergence is never evaluated, so the inference complexity remains $O(ND^2L)$ regardless.

\subsubsection{Problem formulation}

We consider an element-wise nonlinear Gaussian measurement model. The prior over the state $x \in \mathbb{R}^D$ is a diagonal Gaussian $x \sim \mathcal{N}(\mu_0, \mathrm{diag}(\sigma_0^2))$, and the observation is generated as
\begin{equation} \label{eq:nonlinear-obs}
    z = h(x) + \varepsilon, \quad h(x) = x + \alpha \odot x^2, \quad \varepsilon \sim \mathcal{N}(0, \mathrm{diag}(\sigma^2)),
\end{equation}
where $\alpha \in \mathbb{R}^D$ controls the strength of the quadratic nonlinearity and $\odot$ denotes element-wise multiplication. The quadratic twist $h(x)$ creates moderately non-Gaussian posterior marginals. Furthermore, depending on the observation value and nonlinearity strength, the posterior can exhibit bimodality: the equation $\alpha_d x_d^2 + x_d - z_d = 0$ admits two real roots when the discriminant $1 + 4\alpha_d z_d > 0$, creating a secondary posterior mode.

Each inference task is parameterized by $(\mu_0, \sigma_0^2, \alpha, \sigma, z)$. The prior means are drawn from $\mathcal{U}(-0.25, 0.25)$, variances from $\mathcal{U}(1, 5)$, nonlinearity coefficients from $\mathcal{U}(0.1, 0.3)$, and noise standard deviations from $\mathcal{U}(0.5, 1.5)$, all sampled independently per dimension. Because the prior is diagonal, $h(x)$ is element-wise, and the noise is diagonal, the posterior factorizes into $D$ independent 1D marginals, which we exploit to generate ground-truth samples via inverse CDF sampling.

\subsubsection{Experimental setup}

For $D = 10$, the training set consists of 200 tasks, 20 validation tasks, and 100 test tasks. For $D = 15$, we use 500 training tasks, 100 validation tasks, and 100 test tasks. The evaluation uses $10^4$ ground-truth samples per task. The hyperparameters for PINPF and the baselines, determined via ablation on the validation set, are listed in Table~\ref{tab:highd_params}.

\begin{table}[ht]
    \centering
    \caption{Hyperparameter settings for the high-dimensional nonlinear experiments.}
    \label{tab:highd_params}
    \begin{tabular}{@{}lr@{}}
        \toprule
        \textbf{Parameter} & \textbf{Value} \\
        \midrule
        \multicolumn{2}{l}{\textit{PINPF -- MLP Architecture}} \\
        Hidden Layers / Hidden Dim ($D\!=\!10$) & 6 / 64 \\
        Hidden Layers / Hidden Dim ($D\!=\!15$) & 8 / 128 \\
        Activation Function & SiLU \\
        \midrule
        \multicolumn{2}{l}{\textit{PINPF -- Training}} \\
        Training particles & 500 \\
        Pseudo-time step & fix, $\Delta\lambda = 0.01$ \\
        Optimizer / Learning Rate & Adam / 0.008 \\
        Batch size & 128 \\
        \midrule
        \multicolumn{2}{l}{\textit{PINPF -- Inference} and {\textit{Daum--Huang flows}}} \\
        Num.\ particles & 2000 \\
        Pseudo-time step & adaptive, $\Delta L = 0.5$ \\
        \midrule
        \multicolumn{2}{l}{\textit{SVGD}} \\
        Num.\ particles & 500 \\
        Iterations & 50 \\
        Learning Rate & 0.2 \\
        \midrule
        \multicolumn{2}{l}{\textit{Annealed MCMC}} \\
        Num.\ particles & 1000 \\
        Temperature steps & 10 \\
        NUTS iterations per step & 2 \\
        NUTS step size & 0.1 \\
        \bottomrule
    \end{tabular}
\end{table}

We note that SVGD can achieve substantially better accuracy when given a higher computational budget. With 1000 particles and 75 iterations, SVGD reaches an ED of $0.189$ on the $D=10$ problem, comparable to PINPF ($0.204$), but at $3\times$ the runtime ($0.163$s vs.\ $0.053$s). We therefore set the number of SVGD iterations to 50 to maintain a comparable inference time.

\subsubsection{Results and discussion}

\paragraph{Quantitative analysis}

\begin{table}[t]
    \centering
    \caption{Performance on the 10-dimensional nonlinear problem. Mean values are reported over 100 test tasks.}
    \label{tab:nonlinear10d}
    \begin{tabular}{@{}lccc@{}}
        \toprule
        Method & ED & SWD & Time [s] \\
        \midrule
        PINPF (ours)                   & 0.2041 & 0.3228 & 0.0531 \\
        \midrule
        Incompressible flow            & 0.9486 & 0.6837 & 0.5668 \\
        Local Gaussian exact flow      & 0.2683 & 0.7221 & 0.0836 \\
        Mean Gaussian exact flow       & 0.2488 & 0.3840 & 0.0639 \\
        SVGD                           & 0.3394 & 0.4455 & 0.0790 \\
        annealed MCMC                  & 0.0388 & 0.2781 & 1.5966 \\
        \bottomrule
    \end{tabular}
\end{table}

\begin{table}[t]
    \centering
    \caption{Performance on the 15-dimensional nonlinear problem. Mean values are reported over 100 test tasks.}
    \label{tab:nonlinear15d}
    \begin{tabular}{@{}lccc@{}}
        \toprule
        Method & ED & SWD & Time [s] \\
        \midrule
        PINPF (ours)                   & 0.2313 & 0.3188 & 0.0663 \\
        \midrule
        Incompressible flow            & 1.2942 & 0.7243 & 0.5613 \\
        Local Gaussian exact flow      & 0.3331 & 0.6597 & 0.1272 \\
        Mean Gaussian exact flow       & 0.3199 & 0.3814 & 0.0963 \\
        SVGD                           & 0.3637 & 0.4182 & 0.0816 \\
        annealed MCMC                  & 0.0504 & 0.2642 & 1.5995 \\
        \bottomrule
    \end{tabular}
\end{table}

Tables~\ref{tab:nonlinear10d} and \ref{tab:nonlinear15d} summarize the results for $D=10$ and $D=15$, respectively.
PINPF achieves the best ED among all deterministic methods in both settings, outperforming SVGD, and the exact flows (mean and localized).
The latter achieve comparable ED to PINPF due to the moderate nonlinearity of the problem, but they are less accurate in terms of SWD.
The incompressible flow performs poorly in both settings, with ED values $4$--$6\times$ higher than PINPF.

Annealed MCMC remains the most accurate method in terms of ED, but at a computational cost $24$--$30\times$ higher than PINPF. Nevertheless, PINPF provides substantially better accuracy than SVGD and the analytic flows while maintaining the lowest inference time per task.

The scaling from $D=10$ to $D=15$ shows a modest increase in ED ($0.204 \text{ vs. } 0.231$) and a similar result in terms of SWD ($0.323 \text{ vs. } 0.319$), confirming that the method generalizes gracefully to higher dimensions regarding this problem class. The inference time increases by approximately $25\%$, which is consistent with the increased computational cost of evaluating the flow and its gradients in higher dimensions.

\paragraph{Ablation on training step size}

\begin{table}[t]
    \centering
    \caption{Effect of training pseudo-time step $\Delta\lambda$ on inference quality ($D=10$, $N=2000$, $\Delta L=0.5$, 100 test tasks).}
    \label{tab:training_lambda}
    \begin{tabular}{@{}ccc@{}}
        \toprule
        $\Delta\lambda_{\text{train}}$ & ED & SWD \\
        \midrule
        0.002 & 0.2294 & 0.3389 \\
        0.010 & 0.2041 & 0.3228 \\
        0.050 & 0.2403 & 0.3519 \\
        0.100 & 0.6186 & 0.5214 \\
        \bottomrule
    \end{tabular}
\end{table}

Table~\ref{tab:training_lambda} shows the sensitivity of PINPF to the training pseudo-time step $\Delta\lambda$. Since training cost is proportional to the number of Euler steps, smaller $\Delta\lambda$ increases the computational burden. A step size of $\Delta\lambda = 0.01$ (100 steps) achieves the best performance, and reducing it further to $\Delta\lambda = 0.002$ (500 steps) does not improve accuracy despite the $5\times$ increase in training time. Conversely, coarser discretizations degrade performance: $\Delta\lambda = 0.05$ yields a moderate increase in ED, while $\Delta\lambda = 0.1$ leads to severe deterioration (ED increases $3\times$),  indicating that the PDE residual is inadequately resolved at coarse temporal discretizations.

\paragraph{Ablation on flow input features}

\begin{table}[t]
    \centering
    \caption{Feature ablation on the $10\,$D nonlinear model ($N=2000$, $\Delta L=0.5$, 100 test tasks).}
    \label{tab:feature_ablation}
    \begin{tabular}{@{}lcc@{}}
        \toprule
        \shortstack[l]{Additional input features\\beyond $(x, \lambda, z)$} & ED & SWD \\
        \midrule
        none                                   & 0.6181 & 0.5162 \\
        $\log h$                               & 0.5105 & 0.4672 \\
        $\nabla \log p_\lambda$                        & 0.4318 & 0.4328 \\
        $\nabla \log h$                        & 0.2503 & 0.3714 \\
        $\nabla \log g$                        & 0.6466 & 0.5169 \\
        $\nabla \log p_\lambda, \log h, \nabla \log h \text{ (all)}$ & 0.2041 & 0.3228 \\
        \bottomrule
    \end{tabular}
\end{table}

Table~\ref{tab:feature_ablation} presents a detailed decomposition of the feature contributions (cf.\ Eq.~\ref{eq:mlp-features}) on the $10\,$D nonlinear problem.
Among the individual features, the measurement gradient $\nabla \log h$ is the most informative, it achieves an ED of $0.250$ on its own. This is consistent with the theoretical structure: the quadratic nonlinearity makes $\nabla \log h$ a nontrivial function that encodes the local curvature of the likelihood surface. The target score $\nabla \log p_\lambda$ provides moderate improvement, while $\log h$ alone contributes the smallest reduction. The prior gradient $\nabla \log g$ alone offers no benefit, as the prior is a simple diagonal Gaussian whose gradient carries limited information beyond what is already encoded in the baseline input.

The combination of $\nabla \log p_\lambda$, $\log h$, and $\nabla \log h$ achieves the best performance, confirming that these features are complementary and that the feature selection strategy of the input vector \eqref{eq:mlp-features} is well-motivated also in high-dimensional nonlinear settings.

\section{Conclusion} \label{sec:conc}

The paper has presented the physics-informed neural particle flow for the Bayesian update step, which integrates the log-homotopy approach with neural operator-based inference. Coupling the log-homotopy trajectory with the continuity equation yields the master PDE, which serves as a physical constraint on the transport velocity field. This unsupervised training framework does not require ground-truth posterior samples.
PINPF departs from the stochastic relaxation of MCMC and Langevin dynamics in favor of finite-time deterministic transport. Unlike black-box neural operators, embedding the master PDE into the training objective ensures that the learned transport respects probability conservation.

Analysis demonstrates three advantages of this approach. First, the neural parameterization acts as an implicit regularizer, reducing the numerical stiffness inherent in exact analytical flows. This allows larger integration steps and fewer function evaluations. Second, the method maintains linear computational complexity with respect to particle number, making high-dimensional nonlinear filtering tractable. Third, the physics-informed nature of the operator ensures robust zero-shot generalization to out-of-distribution priors, outperforming purely data-driven approaches.

Experiments on multimodal Gaussian mixtures and a non-Gaussian nonlinear scenario confirm that PINPF accurately captures complex posterior geometries and resolves multiple modes. This work proves that neural transport operators constrained by physical conservation laws offer a viable, efficient alternative for Bayesian inference.

Besides the demonstrated advantages, the proposed framework has the following limitations. In the presence of extremely sharp likelihoods, the gradients of the log-likelihood become exceedingly large. This reintroduces severe numerical stiffness into the flow dynamics, which forces the adaptive solver to take very small integration steps, thereby increasing inference time. Second, because the master PDE is enforced as a soft penalty within an empirical loss function, there is no theoretical guarantee that the learned neural operator satisfies the continuity equation exactly everywhere. The resulting transport map remains an approximation bounded by the network's representational capacity and the finite sampling of the training dataset. Third, because the master PDE (and consequently the loss function) explicitly relies on the intermediate score $\nabla \log p_\lambda = \nabla \log g + \lambda \nabla \log h$, the framework fundamentally requires an analytically differentiable prior. Applying this method directly to non-parametric priors (e.g., a raw particle cloud) requires an intermediate score-estimation step, which may introduce approximation errors. Finally, while processing particles independently allows for highly efficient $O(N)$ inference, the network relies entirely on local analytical gradients. Consequently, particles may struggle to route mass efficiently across highly disjoint priors, and the method cannot natively process non-parametric priors. This limitation could be addressed in future work by extending the architecture with Deep Sets or attention mechanisms to condition on the empirical population, exchanging some computational efficiency for global geometric awareness.

\appendix
\section{Training samples} \label{sec:app}
\begin{figure*}[h]
    \centering
    \includegraphics[width=\textwidth]{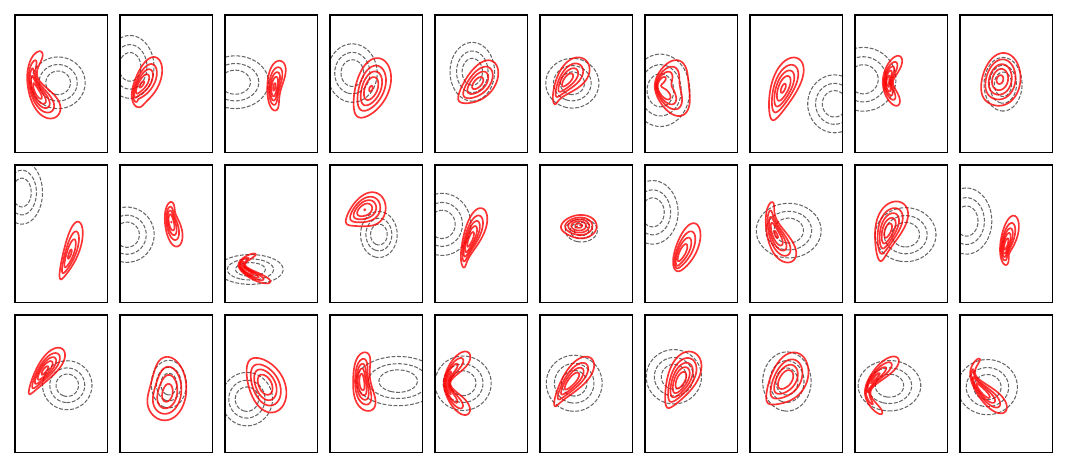}
    \caption{Random samples drawn from the training dataset for the TDOA problem. The dashed black lines represent the Gaussian prior, and the solid red lines represent the true posterior density.}
    \label{fig:training_samples}
\end{figure*}

\begin{figure*}[h]
    \centering
    \includegraphics[width=\textwidth]{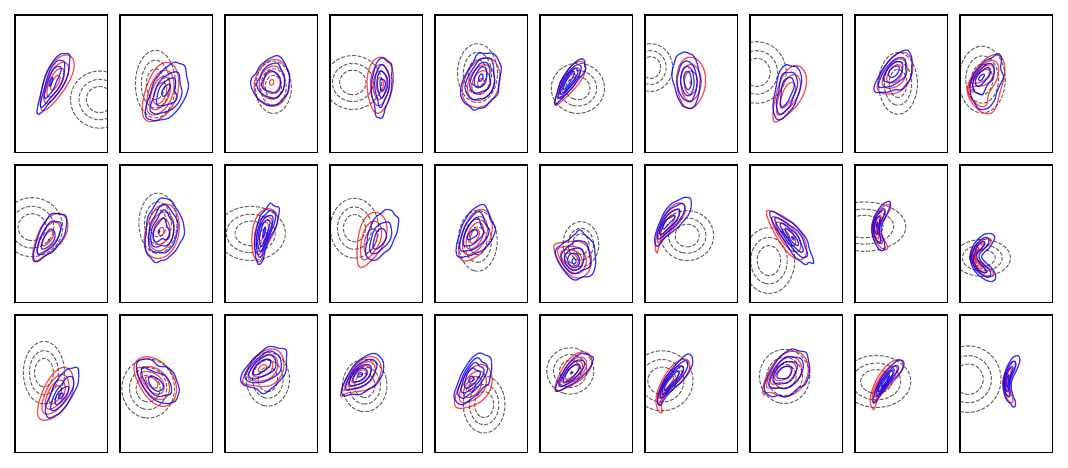}
    \caption{Randomly selected samples from the test set comparing the ground truth posterior and the proposed Neural Flow method. Black dashed: Prior distribution. Red solid: Ground truth posterior. Blue solid: Kernel Density Estimate (KDE) of particles generated by our Neural Flow model.} 
    \label{fig:test_samples_matrix}
\end{figure*}

\begin{figure*}[h]
    \centering
    \includegraphics[width=\textwidth]{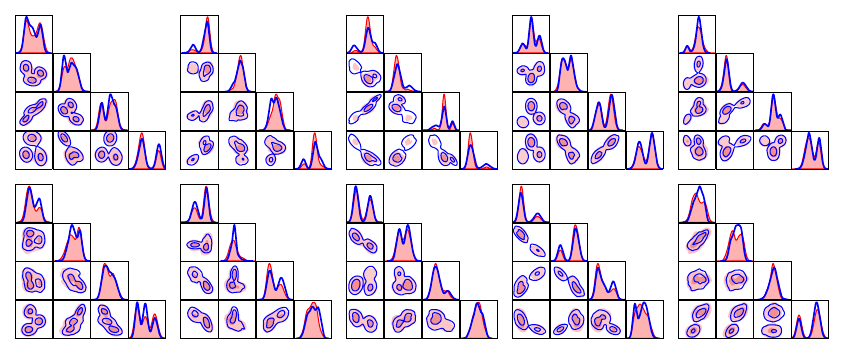}
        \caption{Cornerplots of randomly selected test samples of the four-dimensional Gaussian mixture posterior problem. Red solid: Ground truth posterior. Blue solid: kernel density estimate of particles generated by PINPF.}
    \label{fig:gmm-matrix}
\end{figure*}

\printcredits

\bibliographystyle{elsarticle-num} 

\bibliography{ref}

\end{document}